# A Survey on Trust Metrics for Autonomous Robotic Systems

Vincenzo DiLuoffo, William R.Michalson

*Robotics Engineering*
*Worcester Polytechnic Institute (WPI)*
*100 Institute Rd, Worcester, MA 01609*
*{vdiluoffo, wrm} @wpi.edu*

*Abstract*— **This paper surveys the area of "Trust Metrics" related to security for autonomous robotic systems. As the robotics industry undergoes a transformation from programmed, task oriented, systems to Artificial Intelligence-enabled learning, these autonomous systems become vulnerable to several security risks, making a security assessment of these systems of critical importance. Therefore, our focus is on a holistic approach for assessing system trust which requires incorporating system, hardware, software, cognitive robustness, and supplier level trust metrics into a unified model of trust. We set out to determine if there were already trust metrics that defined such a holistic system approach. While there are extensive writings related to various aspects of robotic systems such as, risk management, safety, security assurance and so on, each source only covered subsets of an overall system and did not consistently incorporate the relevant costs in their metrics. This paper attempts to put this prior work into perspective, and to show how it might be extended to develop useful system-level trust metrics for evaluating complex robotic (and other) systems.**

## I. Introduction

We have surveyed the trust metric space related to evaluating systems, hardware components, software components, cognitive-layer robustness as well as vulnerabilities introduced in the supply chain and have come to realize that no current set of metrics for assessing system trust fully spans any system architecture, let alone autonomous robotic systems. The overall complexity of performing assessment and the complexity of identifying potential security problems are bad enough in a controlled environment; now add high-value targets in an unconstrained environment and they get a lot worse.

Defining trust metrics for system security is difficult [1] leading many practitioners to only define metrics for small portions of an overall system. This approach, while making the assessment of a system more tractable, can result in security vulnerabilities being undetected. Existing approaches to evaluating trust rapidly become computationally intractable due to the large number of interrelated variables that must be considered. Thus, there is a need to come up with a way to make evaluating system's trust more computationally feasible.

We also observed that many approaches tend to focus on a small stovepiped set of problems which are more-or-less tractable. While that body of work provides a useful foundation, it omits several factors that are important to consider when taking a holistic approach to trust evaluation. For example, neither loss (the cost of damage from an exploit) nor reward (the benefit of exploit) costs were addressed for all facets of a system, even though these costs have an effect on the level of trust that can be assigned to a system. In our opinion, there is a need to have a secure base before trust can be extended with external entities by evaluating an internal representation. A secure base is the set of security features (hardware and software) supported by the platform. Once those features are understood the system can internally analyze and evaluate how to react to external stimuli such as external requests received through network interfaces. In essence, this internal evaluation relies on a model of the security features and how those features interact to detect and/or mitigate requests that may have malevolent intent. Such a model requires metrics to rate the level of trust for the best possible security posture. The metrics used in the model help establish the chain of trust depending on the configuration and features of the autonomous robotic system.

Autonomous robotic systems, such as autonomous vehicles operating within a dense population environment, can be very complex. Such a robotic system is typically constructed from many individual components including hardware (CPUs, sensors, actuators, accelerometers, systems of systems), software (firmware, OS, services, cognitive layer, application specific logic), and AI components (learning algorithms). These components may be sourced through one or more supply chain vendors. The integration of all these components becomes a complex problem that significantly effects the trust model.

For example, components of a system may be open source, Commercial Off the Shelf (COTS), or custom built. An individual component may work as designed, but when integrated with other components the behavior of the composite system may be unexpectedly altered in a way that exposes security side channels. We believe that a holistic trust metric that accounts for these complexities is beneficial to determining the security posture of a robotic system.

So, there are two main problems in the art that must be addressed. One is simply defining trust and determining ways of establishing the level of trust that can be attributed to a system and/or components of a system. The other is bounding the computational complexity of a system trust model such that the problem can be solved in a reasonable amount of time with reasonable resources.

There are many definitions of the word "Trust" in the context of system security. One definition that captures the essence of the term is that "trust" is "the firm belief in the competence of an entity to act as expected such that this firm belief is not a fixed value associated with the entity but rather it is subject to the entity's behavior and applies only within a specific context at a given time" Azzedin and Maheswaran [2].

In their earlier work, Castelfranchi and Falcone define the concepts of internal vs external, or global trust, with regard to the cognitive social trust model [3]. They further breakdown internal trust into two areas called reliance and disposition, where the former is the ability, competence, and self-confidence; the latter being the willingness, persistence, and engagement to fulfill a task. The external, or global, part of the trust model is to have the opportunity, for the quickest fulfilment of the resources and the success without having interferences and adversities. Other work validates the concepts of internal vs external trust as an example of work from internal trust Devitt's model [4] and Henshel, et al [5] contribution validates the need to separate trust into two categories.

It is important to differentiate these two types of trust concepts when modeling the behavior of autonomous systems as further decomposition can be made in the security validation and protection architecture. Viewing trust from an internal and external perspective can enable the decision process to be different when analyzing the security capabilities of an autonomous system. A system that has more security capabilities might be better suited for requests being made from an external entity vs one that has less capabilities. This type of internal rational is different from today's computer security models where no introspection is performed on the decision being made from an external request. Today's computer security model is to authenticate first then authorize; once authenticated the process or application is granted access to fulfill the request. The term introspection is the capability to look inside and analyze or debug.

Most literature focuses on external communications from the perspective of an agent interacting with another, or a group of agents. This interaction builds a trust evaluation mechanism based on prior evidence and ratings from external entities, or recommenders. Hearing from multiple sources builds confidence about an outcome or decision that others have made. Therefore, that accumulated evidence can sway a choice or further aide in a choice. Some trust rating systems targeting the use of referrers, or recommenders, are discussed in [6], [7], [8], and [9]. However, it is necessary to be cautious when applying these evaluation techniques to an autonomous system.

Historically, the notion of trust is defined based on what is being trusted. For example, only focusing on the individual system architecture, hardware, or software layers of a system. In general, trust is assessed by comparing features of the thing being assessed to a set of guidelines, and this comparison may be done at a high (abstract) or low (detailed) level. High level assessments tend to be qualitative in nature, while lower assessments tend to *appear* quantitative. However, one problem that arises from performing assessment at low levels is that an ad hoc picking of features or weights can result in even low-level assessment being fundamentally qualitative. Further, applying low level assessment techniques to large portions of a system can become expensive both in terms of engineering time and computational resources.

Our motivation for this work is to construct a security assessment framework that first performs the equivalent functions of static analysis and later have the capability to be extended into the form of dynamic analysis for autonomous robotic systems. Security assessment can be performed at the system level using Common Criteria methodology as one approach, which is a static approach that has many paper/human tasks. Alternatively, software uses automated tools to perform static/dynamic analysis. By mimicking the automated analysis tools in the software model, this can be leveraged for our consideration.

The term static analysis is common in today's software development process since the analysis is performed during development/testing or during build cycles. These static analysis tools support many flavors of development languages, such as C++/C, Java, and Python. They work by scanning the source code for potential bugs or security vulnerabilities against a set of rules. These pattern rules can check for syntax issues, buffer overflow issues and null pointer dereferences to name a few. Some tools can perform the same analysis on object code. Common tools like Coverity and Veracode perform these types of analysis. To extend the static analysis for software, there are dynamic tools that perform similar functions, but at runtime. These dynamic tools look for flow control failures, memory errors and race conditions to name a few. A common tool like Valgrind performs these types of analysis. Taking the concept of both static and dynamic analysis to determine potential security risks on an autonomous robotic system needs to encompass the holistic architecture model, and not just software.

It is our belief that developing a security analysis tool to perform a system-level trust assessment is a way that provides useful results in a tractable manner. A more comprehensive security posture can be created when internal and external interactions are separated, but not decoupled. A self-assessment can be performed on an external request or goal, before being fulfilled. This internal checking may uncover, potential vulnerabilities, misbehavior, or lack of ability to perform a task.

Representing an internal system model; may require a large amount of data and may become computationally complex. A Bayesian Network (BN) is a probabilistic graphical model where nodes and arcs define the casual inference using probabilities. Using a BN is one potential solution, as it satisfies the local Markov property where the joint probability is reduced to a compact form, allows the combining of supporting evidence or negating hypotheses. This helps to reduce the computational complexity issue. We also bound the problem by limiting the number of parent nodes in the model.

The goal of this paper is to survey the current techniques for evaluating trust, and to extend them in a structured way both to account for the complexities of autonomous robotic systems and to minimize subjectivity. This paper is structured as follows: section 2 discusses the search results, the trust metric evaluation for layers system, hardware, software, Cognitive/AI, and supply chain can be found in sections 3, 4, 5, 6, and 7. Section 8 describes the assessment of the techniques used in security assessments and this is followed by a solution outline in section 9. We conclude our findings with developing new trust metrics for an autonomous robotic system.

## II. SEARCH RESULTS

So, how does one evaluate or assess trust? In order to perform assessment, there needs to be a metric, or set of metrics, that correlate well with trust and a tractable methodology to evaluate the metric(s). This has generally been done by assigning weights to features and then combining those weights in some way to develop a score that is presumed to correlate with the level of trust assigned to the system being evaluated.

Trust evaluation can be viewed at different levels for computer systems. At the top is the system level architecture, which then breaks down into other individual categories. For purposes of this discussion we focus on, without limitation, the hardware and software subsystems of the overall system. This structure is shown in Figure 1 where the three circles represent the different levels.

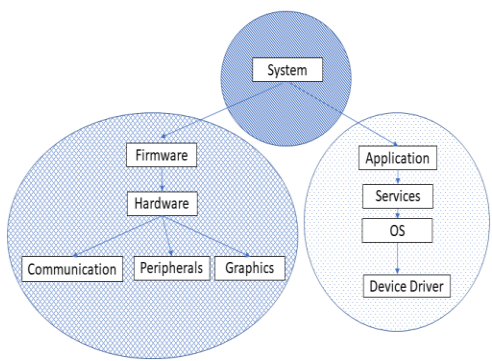

*Figure 1: Different approaches to trust at system, hardware, or software levels*

For evaluating trust in a computer system, the Trusted Computer Evaluation Criteria (TCEC) standard was created in the 80's which disclosed a methodology for evaluating computer systems known as the "Orange Book." The Orange Book defined four primary levels, where A was the highest, followed by B, C and D (being no security). Around the same time another standard was introduced called the Information Technology Security Evaluation Criteria (ITSEC) which defined seven levels (E6, the highest to E0, no security). The levels of ITSEC were mapped to the Orange Book levels that included all the associated sub-levels. The Canadian government created their own version of a security standard and so, to minimize proliferation of standards, the International Standards Organization created the Common Criteria (CC) standard that tried to harmonize these different standards into one, called ISO-15408. The CC has levels ranging from E7 (highest) to E0 (no security). The higher levels of evaluation incorporate formal methods for system behavior verification, auditability of development artifacts and certified 3$^{rd}$ party lab testing.

Using these standards, systems are evaluated based on the architectural features they contain. A system architecture can be evaluated by reading the description of the features required by a trust level and comparing those features to the features of the system being evaluated. Conversely, if a system designer wanted to achieve a desired level of trust, those descriptions provided a list of required features to achieve that level. Since this approach has some level of industry acceptance, it is desirable to capture elements of this method of scoring system-level trust; and expanding on it to capture additional elements determined to be important as well

As is the case at the system-level, similar approaches exist for evaluating trust in the software components of a system. When evaluating software systems, different methodologies have been proposed to ensure a secure software development lifecycle process (SDLC). This means that security tasks are included in the development process and not as an afterthought. These tasks may include code review, vulnerability analysis, white hat testing, and utilizing code scanning software for static and dynamic analysis. Incorporating these techniques in the development process is less expensive than finding security vulnerabilities after release. Some examples of SDLC are the software assurance maturity model; (which is derived from the capability maturity models), the software security framework, the System and Software Integrity Levels from ISO-15026, and the Common Criteria (CC). Following an SDLC process requires tightly integrating the process into the organization's software development philosophy and training their software development resources. Depending on the level that each of these development processes use, a set of tasks are checked off to ensure that they were completed. Software is then evaluated based on these development processes and, as an artifact of the process, the results can be audited.

As we saw at the system level, existing methodologies for trusted software development also rely on formalizing a set of criteria that are applied to the software development process. Again, these criteria form an industry-accepted basis for creating a trust metric that is based on objective criteria.

Not surprisingly, hardware development is taking steps similar to those used in software development to ensure trust in the hardware development process. This is logical since, in modern hardware design, software tools are commonly used to synthesize (or at least customize) hardware components using a variety of hardware design/description languages and other computer aided design tools. Different methodologies are used at this level including the DARPA-led "Trust in Integrated Circuits" program where ICs are manufactured in certified labs to ensure their pedigree, as well as incorporating defined security tasks into the hardware development lifecycle. Indeed, some Electronic Design Automation (EDA) tools are now incorporating security tasks into their tool suites. For example, Mentor Graphics TrustChain provides a unique id that is embedded in the chip and enables tracking in the chip's lifecycle. These methodologies provide an evaluation mechanism for assessing how/if security considerations are being incorporated into the hardware lifecycle.

Thus, the existing art provides some examples for which trust evaluation occurs at the system, hardware, or software levels. While these previous approaches focus on evaluating on a specific level, this may create gaps in the overall security posture of a system. For example, GreenHills INTEGRITY separation kernel was evaluated at EAL 6+ (only software was evaluated) but included a configuration with an embedded PowerPC and PCI card. This evaluation only examined a small portion of an operating system that did not include other services like device drivers or application logic that would be part of a system architecture. Nor was hardware included that might expose side channels into the kernel itself. Traditionally, we have focused the evaluation of trust on just the hardware and software levels of the computer system, but with new autonomous robotic systems, this needs to change.

Autonomous robotic systems are different from computer systems in general; since they may have multiple compute nodes, and are a system of systems, which may include sensors, actuators, and AI operating in unconstrained environments. So, what is a holistic approach to defining a set of security metrics for evaluation? It is our recommendation that incorporating metrics for the system, hardware, software, AI, and supplier, defines a good basis for evaluation. Further, by developing a computationally tractable methodology for trust evaluation that includes metrics spanning all these systems, we form a foundation that can be extended to include additional system aspects not discussed herein.

Figure 2 shows an example of an autonomous robotic system broken down into components. The left side of the figure represents the software layers including the cognitive or Artificial Intelligence (AI) layer, which is expressly identified as a component of the system. The right side represents the hardware layers and the "supply chain", which may influence any part of the system. Components like sensors and actuators are linked to IoT devices since these devices have similar trust issues. Several components did not have a direct mapping to a specific category and were placed into a major grouping (system, hardware, software, AI robustness or supply chain).

For each system component, we searched relevant terms to generally identify the existing art using terms like BNs, trust metrics, risk assessment, and vulnerability assessment for autonomous robotic systems. Other terms were added for each system specific aspect. We then categorized that art according to its primary contribution(s) to modeling the system architecture aspect that were conducive to our goals. The primary contributions were further analyzed to extract the main points that aligned to our objectives, while discarding unrelated findings in a second pass. The litmus tests that we targeted were related to the method(s) for utilizing trust metrics; how were the metrics created, were the metrics linked to standards or recognized bodies of work, and did the findings prove successful. Those that passed; were incorporated into this document, which are further analyzed in section III.

We began our search for trust metrics by attempting to identify those methodologies currently existing in the art that have already been proposed for evaluating trust, even if those methodologies only solved a portion of what we see as the overall problem. In that investigation, we specifically looked for trust evaluation methodologies applied at the system level, as well as methodologies applied at the hardware, software, cognitive/AI, and supply chain levels of a system design.

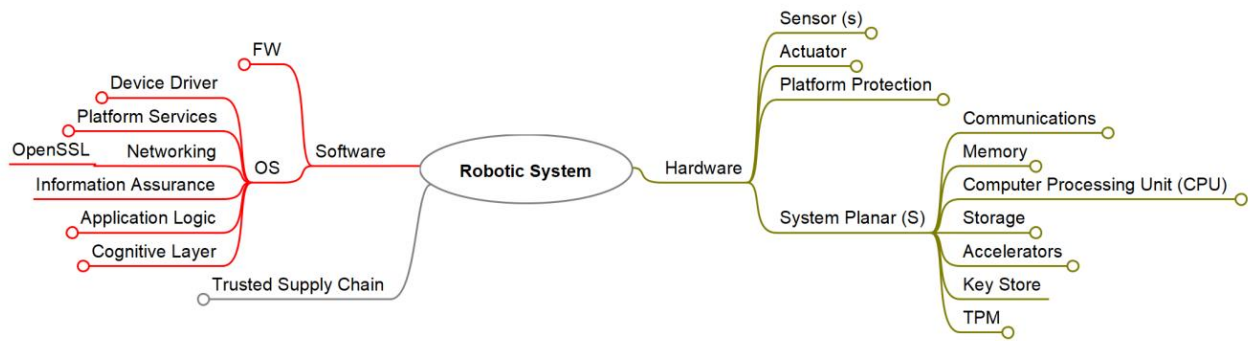

*Figure 2: Mapping search results to each robot component, for trust metrics*

## III. SYSTEM LEVEL TRUST EVALUATION

To identify methodologies used for evaluating trust at the system level, we searched for papers focused on trust metrics, autonomous systems, BNs, trustworthiness, system security, resilient systems, and security evaluation. From these parameters the results are broken down into four groupings: risk assessment, trust with a human in the loop [10], a trusted boot scheme [11], and standards (that included frameworks and security evaluation methodologies [12], [13], [14], [15], [16] ).

We summarize some of the system search results that are attractive to incorporate into our holistic security model. These findings include the need to: support attack paths; determine behavioral states of an autonomous system, and integrate the concepts of trust, resilience, and agility. The system model should also incorporate attributes for physical protection and safety, since these elements will be needed for autonomous robotic systems that are exposed to and interact with humans. In the remainder of this section, we briefly describe what these various techniques are, how they work, what they do well relative to our goals, and any shortcomings of the technique when applied to autonomous robotic systems.

In our research we found that techniques fell loosely into two categories: graph-based assessment techniques that incorporated metrics and system security assessment techniques. Exemplary graph-based techniques are found in the work of Shetty [17], Henshel et al [18], and Cho et al [19].

For example, Shetty presents a technique to categorize attack paths (a chain of exploits) utilizing a BN where exploit impact, cost of exploit and the degree of difficulty is determined [17]. An attack graph is a visual representation of potential paths that an attacker can take or has taken to achieve a successful goal. It is important to autonomous robotics systems because defense strategies can be used to block the threat ahead of a potential attack.

**What is it** - Shetty's Cyber RIsk Scoring and Mitigation (CRISM) tool is described as interfacing with the Common Vulnerability Scoring System (CVSS) and the National Vulnerability Database (NVD) to create trust metrics. These metrics are derived from the CVSS base score as low (0 to 4), middle (4 to 7), and high (7 to 10). The base score reflects the severity of a vulnerability based on its intrinsic characteristics like the exploitability (attack vector) or the impact to the data's confidentiality, integrity, and availability. A further explanation is given in the software section below.

**How does it work** – As a first step, an attack graph is generated from network scans, internal enterprise vulnerability tests, internal enterprise vulnerability database (s) for how the systems are connected and network topology. Essentially, this step is to create the topology of how components are connected in the network-based system architecture. The National Vulnerability Database (NVD) is a repository which provides CVSS base metrics on Common Vulnerabilities and Exposure (CVE) entries. Once a vulnerability is discovered, it is assigned a unique CVE Identifier. For example, CVE-2014-7173, a brief description (FarLinX X25 Gateway through 2014-09-25 allows command injection via shell metacharacters to the files sysSaveMonitorData.php, fsx25MonProxy.php, syseditdate.php, iframeupload.php, or sysRestoreX25Cplt.php) and a score of 9.8. The score is derived from the CVSS base metric, including the exploitability (attack vector) and impact (confidentiality, integrity, availability). Only system component vulnerabilities that are associated with the attack graph are reported. NVD assigns metrics that are based on CVSS base scores (low 0 to 4), middle (4 to 7), and high (7 to 10) but only the impact scores are used. A risk probability is calculated depending on the lifecycle of the vulnerability (not yet discovered, discovered, patch available, patch not applied, patch applied, etc.). By combining the risk probability and CVSS impact score, a set of values can be constructed using the information from the attack graph/topology where component security metrics are categorized by attack paths with attributes that include the impact, cost, and degree of difficulty. A risk assessment tool is created that captures the attacker's exploits.

**What does it do well** – From CVSS base metrics (see Software Metric section) NVD assigns a ranking low, middle, and high for vulnerability risks. This provides an established set of data sources and metrics for known attack vectors and paths taken, which are then assessed using a BN. This concept of risk metrics can be extended into different layers of a robotic system using BNs.

**Shortcomings** - CVSS/NVD provides a good framework for systems where the attack vectors are known, but since autonomous systems are relatively new, the attack vectors are not yet completely known. Yet, many parts of a current robotic system can already use this information to at least assess parts of a system. Furthermore, CRISM provides a framework for including these attacks as they are discovered.

Henshel et al, presented a cyber security risk assessment model, that characterized "Trust" as human (users, defenders, and attackers) behavior and all other as "Confidence" (hardware and software).

**What is it** - Trust in humans had two categories: the first being inherent (part of the individual that is further broken into behavioral and knowledge/skill characteristics); and the second being situational (external to the individual). These two categories define how mental states affect risk and impacts the levels of trust [5]. The human factor was being incorporated into a cybersecurity risk framework/model called Multi-Level Risk Assessment Parameterization Framework [18] using BNs.

**How does it work** - The risk assessment taxonomic parameterization framework (i.e. MulRAP Framework) is applied by first identifying the complex system in question. The complex system is then deconstructed into its functional components and processes. The level of deconstruction is determined by the granular specificity of the risk assessment question. The functional components and processes (i.e. system/risk parameters) are then characterized based on the environmental context of the risk assessment question and the known vulnerabilities of the complex system in question. A BN allows the factors that contribute toward high-risk situations to be identified. These risks are identified as type of activity from a country, threat index for the country, risk prior to defense, risk after defense, potential access, and network component compromised. Two scenarios (risk of a database being compromised, one with low to medium risk level and the other with high) were given that demonstrated the outcomes from the BN.

**What does it do well** – Henshel et al, does a good job of identifying the various elements that go into evaluating trust assessments from human factors and incorporating them into an evidence-based model. It combines information from human factors, and it provides a model for capturing evidence-based causal relationships between elements.

**Shortcomings** – The focus of the paper was on the parameters for defining human characteristic in a cybersecurity model. The framework/model was discussed in another paper as referenced above, but while a direct mapping between humans and machines (human characteristics to machine characteristics) may be invalid, the methodology applied for assessing trust is beneficial. By breaking down human trust further between inherent and situational characteristics, this provides the basis for reducing risk.

Cho et al, presents a technique using Petri Nets that covers trust, resilience, and agility for multi-domain environments called TRAM. The focus is mainly toward human and machine conflict.

**What is it** - TRAM consists of trust (security and dependability), resilience (fault-tolerance, recoverability, and reconfigurability) and agility (Service Level Agreements (SLA)) elements for measuring system quality in a multi-domain environment. This multi-domain environment consists of hardware, software, network, human factors, physical environments, and the effects of behavior at a component or system level for an overall service.

**How does it work** - A Perti network diagram in Figure 3 shows where metric values, assessments, and threats are links of relationships. Tw refers to trustworthiness of a system with Tw= ($T_r$, T, $R_s$, A) where $T_r$ is the degree of threat, T is perceived trust, $R_s$ is resilience, A is agility of the system and metrics range from [0,1] Cho et al, [19].

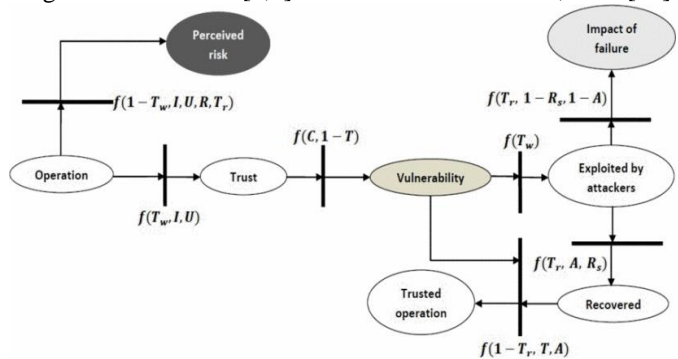

*Figure 3: A Petri net representation of the relationships between metric attributes, assessments, and threats.*

**What does it do well** – The TRAM model ties the relationship of Trust to other variables like resiliency and agility. Autonomous robotic systems will need to support the concept of resilience and agility, since they may defend themselves from threats and still be able to function in a limited capacity.

**Shortcomings** – – While the paper claimed that the TRAM system included hardware and software assessment, no further definition was included nor was there an explanation about where these modules came from. In autonomous robotics systems, hardware can refer to a number of things including compute modules, sensors, and actuators. These are not simple extensions to a conventional computer system connected to a network. While regular Petri network diagrams can be applied to threat models, they do not provide a mechanism for evidence-based reasoning about uncertainty. Informed decisions or reasoning can take place when data is available to support a hypothesis; this is called Evidence-based reasoning, but reasoning about uncertainty is represented as probabilities. Therefore, the concept of linking the relationships between actors can be applied in a BN to provide a better security assessment in the presence of uncertainties.

System security assessment assures that the security requirements are met in the implemented system. The techniques for the assessment can range from a checklist of tasks to having a formal verification process driven by an independent lab. From our research we found that commonly used system security assessment techniques are Federal Information Processing Standard FIPS 140-3 and Common Criteria (CC). The additional benefit of CC is the mapping to safety standards, as discussed below.

An underlying assumption in many of the sources we reviewed was that computing devices are assumed to be housed within a controlled physical location and managed. Since autonomous robotic systems are not in such an environment, and may be exposed to physical attacks (invasive attacks involving physical manipulations on semiconductors like microprobing, and non-invasive attacks where side channel leakage can occur like power analysis) [20], this assumption is invalid and this type of system will need system level protection. System level protection can range from the highest level of security, tamper protection with countermeasures, to lowest level, evidence of tampering. Two specifications that do address the higher level of protection requirements are FIPS and CC.

**What is FIPS?** – The Federal Information Processing Standard FIPS 140-3 is a standard (levels 1 to 4, level 4 being the highest) for approving cryptographic modules, but the development lifecycle methodology for achieving levels 3 and 4 are complex (fully documented and formal design including testing) and require fully vetted modules that undergo security analysis by independent labs. FIPS Level 1 is the minimum set of security requirements that uses at least one approved algorithm or security feature. Level 2 builds on one by adding the tamper evidence requirement, such as a seal or coating that when broken provides visual evidence of tamper. Security Level 3 is intended to have a high probability of detecting and responding to attempts at physical access, use or modification of the cryptographic module. Security Level 4 provides the highest level of security, the physical security mechanisms provide a complete envelope of protection around the cryptographic module with the intent of detecting and responding to all unauthorized attempts at physical access. Security Level 4 cryptographic modules are useful for operation in physically unprotected environments.

**How does FIPS work** – FIPS validation requires fully vetted modules that undergo security analysis by independent labs and the implementation of algorithms are tested against a known set of algorithm values. At higher levels, the certifying module undergoes a series of exploits by the lab to ensure that there are safeguards in the system that are present and that these safeguards work against the expected attack vectors. Attack vectors can range from environmental stress testing including temperature (low and above operational levels) and physical testing, which attempts to break into the tamper protected envelope by any means. If the module can successfully defend against these different scenarios, it is deemed as passing. The cost and time frame to achieve certification depends on a number of factors and of course the level being tested against. The timeframe can take from 6 to 24 months. Government recovery costs range from $4k for level $1 to 11k for level 4, but does not include lab, consultant, and own internal team and development costs. A process flow for FIPS begins with block 1, in a five-block serial process flow. In block 1, the implementation under test undergoes activities like contract with lab, identify what to validate, perform a gap analysis (FIPS requirements and current product), fix, update and extend to support FIPS algorithms, prepare documentation for certification, and prepare for testing can take place. Block 1 can take from 3 to 18 months. In Block 2, the request is submitted for review pending and this can take up to six weeks followed by Block 3. In Block 3, a review is performed where questions and answers can be exchanged for a duration of two to three weeks. Block 4 is the coordination period where corrections and revision of documents take place, this period can take from 1 to 6 months. Block 5 is the approval, which can take a week for the certificate.

**What does FIPS do well** – FIPS 140-3 standard provides a methodology for security evaluation depending on the level of features/functions that a module supports. This is a well-documented process with a number of certified labs that support the evaluation. The algorithms also support a set of known good results that are used during the implementation and testing phases. This methodology can be extended in the autonomous robotic system space for covering security modules and tamper protection.

**Shortcomings of FIPS** – This relies completely on the human-based vetting process which is time consuming and expensive. Another shortcoming is that the criteria are static, once assessed, always assessed. There is no way to account for "unexpected" attacks or for having the system reason about what may or may not be an attack. A final point is that the standard defines a set of requirements to achieve a level of evaluation that only targets the cryptographic modules and the small Target of Evaluation (TOE), but this may miss higher levels of side channels.

Like FIPS, the Common Criteria (CC) provides a framework for information technology security evaluation and defines a common evaluation methodology to achieve an international recognized certification.

**What is it** – CC defines the Evaluation Assurance Levels (EAL) are from 1 to 7, with EAL 7 being the highest level. EAL 1 (functionally tested), EAL 2 (structurally tested), EAL 3 (methodically tested and reviewed), EAL 4 (methodically designed, tested, and reviewed. EAL 5 (semiformal designed and tested), EAL 6 (semiformal verified design and tested. Formal methods and systematic covert channel analysis required.), EAL 7 (formally verified design and tested). This level requires more formal methods and systematic covert channel analysis required [21]. Level 4 is the highest level that is mutually recognized by the Common Criteria Recognition Arrangement (CCRA).

**How does it work** – CC defines a process flow for deliverables and certification. A first step in the process flow is to define the security product and market segment for the product. This will be matched to a Protection Profiles (PP) that is used for the type of product or requirements for the needed security solution, which is like a request for proposal. The next step is to define a Target of Evaluation (TOE). This defines the security features that the product supports. A protection profile is created to document what the TOE is for the certification. Defining the Security Target (ST) is the next step, which defines the security properties of the TOE like functionality and assurance components; the ST can target multiple PPs. The ST and implementation documentation is provided to an accredited third-party lab. The lab evaluates if the ST meets the PP through testing and makes the decision for the evaluation. The lab performs two evaluation tests called Security Functional Requirements (SFRs) and Security Assurance Requirements (SARs). ST establishes the SFR for the product evaluation depending on the individual security features and the SARs for assurance claims. The SAR describes the measures taken during development and evaluation to ensure that the security functionality claims comply. The lab tests determine if the claims being made are valid. The final step is based on the test findings where the lab assigns an evaluation assurance level. In the case of higher assurance levels formal evidence must be submitted with the documentation. Depending on the EAL, the time and cost can range considerably. Dale, in 2006 presented a briefing that provided the following correlation between the levels: EAL 1 = 0, EAL 2 = cost $100 to $170k and 4 to 6 months, EAL 3= $130 to $225K and 6 to 9 months, EAL 4 = $175 to $750k and 7 to 24 months, EAL 5 = $750 to $2M and 24 to 48 months [21].

**What does it do well** – CC has been recognized internationally as a (ISO -15408) standard for information technology. This standard can be applied to a number of different IT devices, which helps with autonomous robotic systems. A fully documented development lifecycle is required, and at higher assurance levels will undergo an analysis (protection mechanisms) from an independent 3[rd] party certified lab. Other standards for safety have been mapped to these assurance levels, as is discussed below.

**Shortcomings-** Currently CC is being utilized toward IT equipment and will need to include complex systems like autonomous robotics systems where not only the OS will be considered, but sensors, actuators and AI must be part of the certification process. Similar to FIPS, CC may only define a small TOE part of the system and potentially expose other parts of the system. Also, like FIPS, this captures a static assessment of the TOE and does not account for uncertainty.

Both FIPS and CC have evolved from a need to assess cryptographic modules and are early examples of approaches for scoring certain systems and system components. Another linkage is the Orange Book assessment guide for higher levels systems.

FIPS is geared toward cryptographic modules but addresses the tamper protection mechanisms. CC, on the other hand, is targeted toward system level assurance where tamper protection is included in the higher assurance levels. There is a continuous effort in mapping safety and security specifications, since each have similarities in the development process, but more importantly security is becoming a requirement. A general specification on, IEC 61508, spans into the medical, machinery, automotive, rail, process industry and nuclear domains. The European Union is taking steps toward creating workgroups to address the mapping of safety and security [22]. Like FIPS and CC, such specifications aid in quantitatively assessing systems and components, as they provide a basis for scoring systems and components.

Since autonomous robotic system can be incorporated into a number of industries the current safety standards are used to mapping back to a CC standard. Schmittner and Ma, presented a paper that discusses the mapping of Automotive Safety Integrity Level (ASIL) to CC EAL levels [23] as shown in Table 1 .

*Table 1: Comparison of integrity and assurance levels*

| ASIL   |   | EAL   |
|--------|---|-------|
| ASIL A | ~ | EAL3  |
| ASIL B | ~ | EAL4  |
| ASIL C | ~ | EAL5  |
| ASIL D | ~ | EAL 6 |

Table 2 shows the layering of the different safety specifications (automotive, general grouping, aviation and rail) and the mapping to Common Criteria for security [23], [24], [25]. At the top row is the automotive standard ASIL that Schmittner and Ma mapped to CC as shown in Table 1, where Quality Management (QM) represents that risk is not unreasonable and no safety measure is needed. The mapping also had ASIL-D being the highest degree of hazard injury and highest degree of rigor applied in the assurance. Mapping ASIL to General and Rail safety standard, Safety Integrity Level (SIL) where SIL 1 is the lowest and SIL 4 being highest. These safety standards are then mapped to aviation standard D0-178, where Design Assurance Level (DAL) implies A(Catastrophic), B (Hazardous and Sever -Major), C (Major), D (Minor), and E (No Effect). These safety standards are aligned with FIPS 140-3 and CC EALs. CC supports in between levels by adding the plus symbol, which means that partial categories were fulfilled for a specific category. All these techniques are, at root, doing the same thing. The problem is that they do not really capture the details of how a system actually operates (complexity/granularity rapidly becomes intractable using this methodology). The problem is that the assessment is a bit divorced from reality for an autonomous system. If it is available, it is beneficial, but it will not be available for everything. Using it as evidence supporting trust makes sense, but it is only part of the solution.

*Table 2: Mapping Safety and Security specifications*

| Domain | Domain Specific Assurance Levels | | | | | |
|---|---|---|---|---|---|---|
| Automotive (ISO 26262) | QM | ASIL-A | ASIL-B | ASIL-C | ASIL-D | ASIL-+ |
| General (IEC-61508) | - | SIL-1 | SIL-2 | SIL-3 | SIL-4 | |
| Aviation (DO-178/254) | DAL-E | DAL-D | DAL-C | DAL-B | DAL-A | |
| Railway (CENELEC 50126/128/129) | - | SIL1 | SIL2 | SIL3 | SIL4 | |
| FIPS 140-3 | L1 | L2 | L3 | → | L4 | → | → |
| CC (ISO 15408) | EAL1 | EAL2 | EAL3 | EAL4 | EAL5 | EAL6 | EAL7 |

The main value in these approaches is their identification of areas that need to be assessed in a comprehensive model for a robotic system, and their proposed methodology for assigning qualitative "weights" based on component (system) characteristics. Using CC EAL levels 1 to 5 can be looked at for a base set of system level metrics to cover the security and potential safety assurance. We also need to consider resiliency and how the system will react when threats are encountered. Cho et al, discussed resiliency in their TRAM research as these concepts should be considered.

IV. HARDWARE LEVEL TRUST EVALUATION

A robot has many hardware components from processing data to sensing the environment. The processing of data is performed by a microprocessor and other accelerator devices. Sensors and actuators perform the locomotion, manipulation, and navigation functions in a robot. This integration between the hardware components can be defined as direct or decoupled communications. This means that in the case of direct communications, the data is exchanged between client and server with no blockers, so that a response is provided as a direct means of the request, which is similar to a command-response manner. In the case of decoupling communications, there is a broker as an intermediate step to exchange data and for which the widely used protocol is a publisher/subscriber format. The broker is the man in the middle that needs to know all the entities in the publisher/subscriber network in order to route the messages between source and destination. This might impact trust differently as compared to the direct communications approach. Sensors provide the basic capabilities for autonomous systems to collect data from the environment in order to make decisions. Each hardware component is constructed from a single Integrated Circuit (IC) or a set of ICs that provide the functionality for data processing, sensing or actuator controller logic. ICs are made up of electronic circuits like, transistors, capacitors resistors, etc., that make up the digital or analog functional logic. ICs that are designed for microprocessor, graphics engines, digital processing logic can have many millions of transistors. In order to design these complex ICs software is used to design, develop and test called Electronic Design Automation (EDA) tools. EDA tools provides a number of functionalities like logic synthesis (converts Verilog or Very High Speed Integrated Circuit Hardware Description Language (VHDL) to netlists), place and route (uses netlist to find optimal locations for wires and transistors), simulation (takes the netlist as a description of the circuit logic and mimics its behavior) and verification (design rules for the netlists). So why is it important to have hardware trust?

In order to establish a secure foundational base for autonomous robotics systems a concept called the hardware root of trust is used to form a trusted base for all secure operations. Hardware root of trust is the concept to boot from an immutable point in the software stack used on computer systems today. This concept assumes that the hardware is free from potential vulnerabilities, but from our point of view hardware trust should also look back at the design origin of the ICs. This level of trust needs to be considered for the holistic security model, since underlying vulnerabilities like trojans and malware can be introduced in the IC design phase.

A robot's hardware is likely a combination of COTS and custom electronics from a variety of sources. The components on these electronic assemblies, in turn, come from a variety of sources. We generally assume that the things we buy do what they are supposed to do – i.e., an 802.11 chip does 802.11 wireless communications. If it does, we are happy. However, when evaluating the amount of trust that can be placed in a system, we also need to acknowledge that we have no idea what else it might do, and if any other possible functions may result in system vulnerabilities. Security at the chip level is taken into consideration and finding a set of trust metrics is one component to a holistic security architecture.

Each of these hardware components is constructed of Integrated Circuits (ICs) that may have single purpose or combined purpose where multiple functions are placed on a single die. An example of this structure is shown in Figure 4 which illustrates an example of a common robot sensor, called an inertial measurement unit (IMU). This sensor component includes an ARM-based CPU as well as a motion processing unit, or MPU chip. From the block diagram of the MPU, shown on the left of Figure 4, we see that the MPU is itself a complex system in addition to the ARM CPU. The arrow points to an exploded view of the internal design (consisting of two dies integrated into a single package). One die houses a 3-Axis gyroscope and a 3-Axis accelerometer. Another die houses a 3-Axis magnetometer, "the MPU-9250 is a 9-axis Motion Tracking device that combines a 3-axis gyroscope, 3-axis accelerometer, 3-axis magnetometer and a Digital Motion Processor™ (DMP) all in a small 3x3x1mm package" [26] [27]. Thus, even in a simple sensor there may be a large amount of complexity in which there are several places where trust can be compromised even in a $35 COTS part.

The process to create an IC is very complex, starting with Electrotonic Design Automation (EDA) tools and ending in a foundry. Figure 5 [28] shows the top line as the design flow and the lower line being potential points in the process to alter or inject trojans in the design. Counterfeiting is the illegal forgery of an original part. While it poses an economic threat to the legitimate owner of the Intellectual Property (IP), and possibly a reliability threat for the component, a true forgery should not be any different electrically than the original part. However, since the authorized manufacturer of the part being counterfeited has no control over the counterfeit, this provides a counterfeiter an opportunity to insert logic that can compromise trust. These are only a couple of possible threats that can occur at different points in the IC lifecycle.

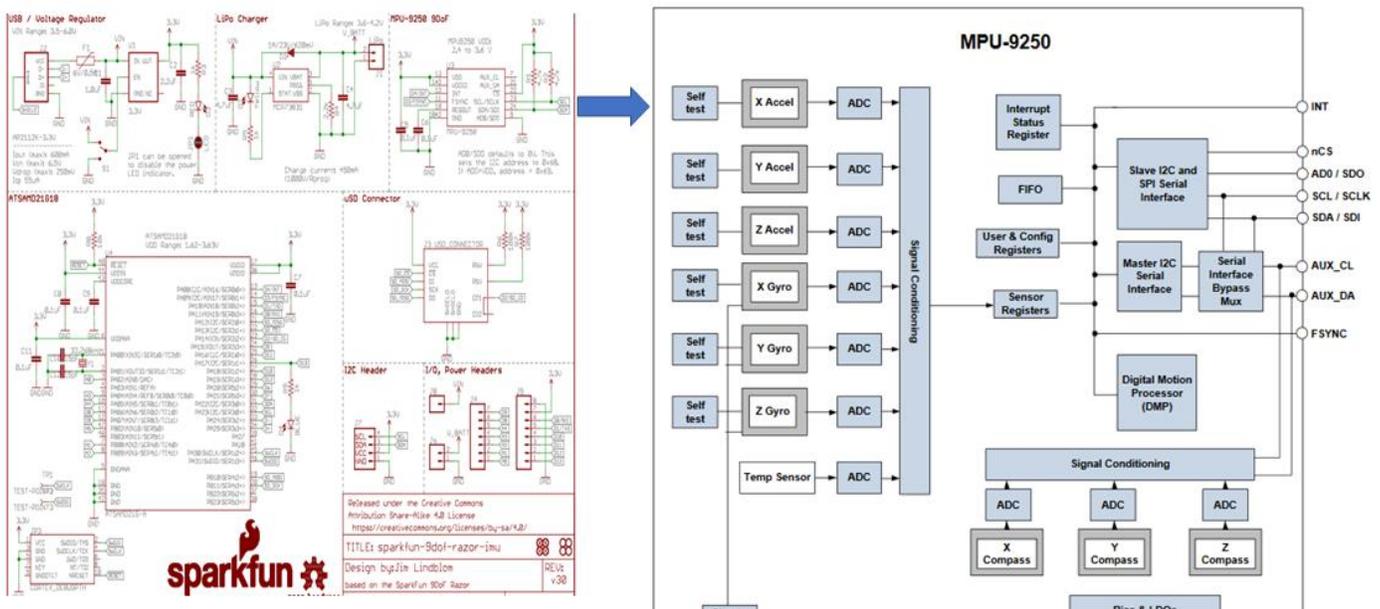

*Figure 4: IMU board with a motion processing unit*

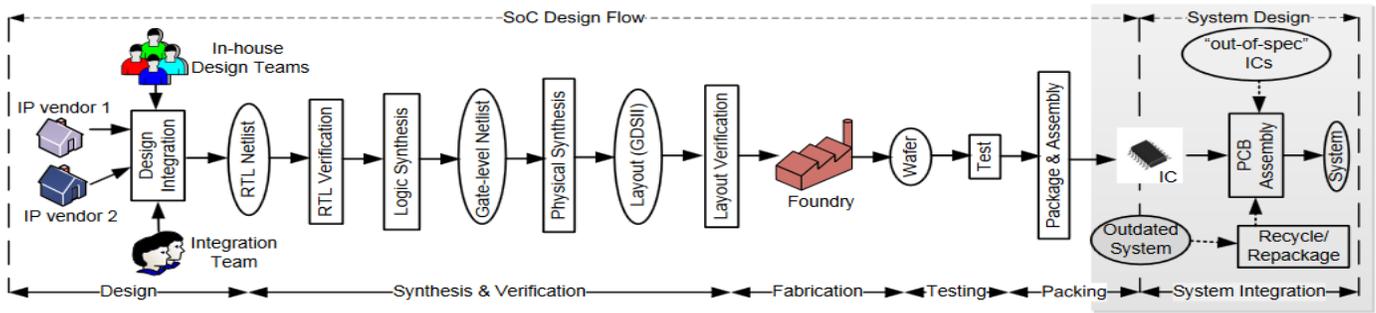

*Figure 5: The top line is the IC design flow, and the bottom line represents the injection points.*

Acknowledging that there are potential vulnerabilities in the design and manufacturing process, there have been a number of initiatives related to trusted IC development including the Trusted Foundry—the proposed approach is to split the foundry functions into two [29]. One foundry would create the fine layer of transistors with detailed leads to connect them. and the second vetted foundry would do the less fine wiring with connections to the outside world. This split benefits trust in two ways, 1) not having a single point of knowledge and 2) it creates a second validation point. We view the foundry as a supply chain trust metric and focus on the IC design in this section.

To identify methodologies used for evaluating trust at the hardware level, we searched for papers focused on BNs, trustworthiness, sensor trust, and IC trust. From these parameters the results were categorized into two groupings, the first being the IoT and Sensor where Mozzaquatro el al [30] and Thamburu el al [31] discuss IoT cybersecurity framework and management system for them. In addition to the first grouping is the work from Lu and Cheng[32] and Boudriga [33] that focused on identifying bad sensor behavior in a wireless network. The second grouping was related to hardware trust metrics, where Kimura presents a technique for determining a design integrity trust metric for hardware design by evaluating five different characteristics ( signal, logical, power, function and structural integrity)[17]. We focus on Kimura's work, since this supports our discussion above with vulnerabilities in the design phase. This is important for autonomous system because hidden threats can be exposed at the design phase and in the supply chain section below, we cover the foundry portion of the IC process as a set of split metrics.

**What is it-** Kimura presents quantifying metrics for hardware designs called "Development of Trust Metrics for Quantifying Design Integrity and Error Implementation Cost". An evaluation of design integrity is accomplished by looking at five different characteristic domains (Logical Equivalence, Signal Activity Rate, Structural Architecture, Functional Correctness, and Power Consumption) of the design and then aggregating their measured deviations from expected characteristics together to arrive at a single value Design Integrity (DI) metric. This technique can be leveraged in two ways, one being a metric for design integrity used by the IC provider to ensure that no unauthorized functions were introduced when the device progressed through foundry and also allows validating the output from the foundry. A consumer of the IC can use the design integrity metrics to determine the differences between actual and expected designs by obtaining the needed information. In each case the Design Integrity approach provides a metric to measure the quality of the design and a methodology to reduce potential exposure to IC threats. An Error Implementation Cost (EIC) measure is developed as a technique to quantify errors and to allow error ranking and rating. A final Trust Measure metric is calculated that includes an estimate of design's integrity and reference design quality or characteristics. In the case of the IC provider, they would have all have content as an output of the product lifecycle and in the case of the IC consumer, they would have limited information, so the reference quality is a variable to address the subcategory content.

**How does it work-** Figure 6 [34] shows the breakdown into the sub layers of the design. The Design Integrity (DI) trust metric accounts for the signal activity rate, logical equivalence, power consumption, functional correctness and structural analysis [35]. The integrity of a design can be defined as the amount of deviation observed between reference and sample designs. For the case of black box IP, a reference specification is usually provided, such that a behavioral model can be constructed, or layout reverse engineering conversion tools can be used to derive netlists.

- Signal activity rate is the number of times the evaluated element changes state over the duration of a given test scheme. Signal rate evaluates data, I/O and logic signals using equation 1.

$$SR_{integrity} = \frac{SR_{expected} - \Delta SR_{dist}}{SR_{expected}} \quad \text{eq (1)}$$

where $0 \leq SR_{integrity} \leq 1$ and the differences between actual and expected is the $\Delta SR_{dist}$ value. This can be used by either the provider or consumer of the IC.

- Logical equivalence is the degree to which the logic state points of the design can be compared to the original reference. This is best suited for the IC consumer case where they may have access to a netlist to ensure that they have a clean chip. A tool like Cadence Conformal can be used to check the difference between netlists and depending on the differences result, this is the deviation score. $LE_{integrity}$ can now be expressed as the ratio of Equivalent Points to the Total Comparison Points as shown in equation 2.

$$LE_{integrity} = \frac{Points_{EQ}}{Points_{COMPARED}} \quad eq\ (2)$$

where $0 \leq LE_{integrity} \leq 1$

- Power consumption measures how closely the actual design aligns to the original reference from a power perspective. Power consumption is comparing the simulation tests to actual tests for each power test point using equation 3.

$$P_{integrity} = \frac{P_{expected} - \Delta P_{dist}}{P_{expected}} \quad eq\ (3)$$

where that $0 \leq P_{integrity} \leq 1$ and the differences between actual and expected is the $\Delta Pdist$ value. This can be used by either the provider or consumer of the IC.

- Functional correctness is the difference between the actual and expected results in order to verify correct design. Tests can range from exhaustive testing to ones that only provide corner and basic coverage. $F_{integrity}$, is evaluated by observing the number of errors that occur, $\varepsilon observed$, for a given verification test scheme and *TPtotal* is the total verification test points used for verifying the design functionality, this is shown in equation 4.

$$F_{integrity} = \frac{TPtotal - \varepsilon observed}{TPtotal} \quad eq\ (4)$$

where $0 \leq F_{integrity} \leq 1$. This can be used by either the provider or consumer of the IC.

- Structural analysis looks at the architectural components regarding gate level net lists for comparison and/or leaf cells using equation 5,

$$S_{integrity} = 1 - \Delta S \quad eq\ (5)$$

where $\Delta S$ is the differences between actual and expected number of modifications found. This can be used by either the provider or consumer of the IC.

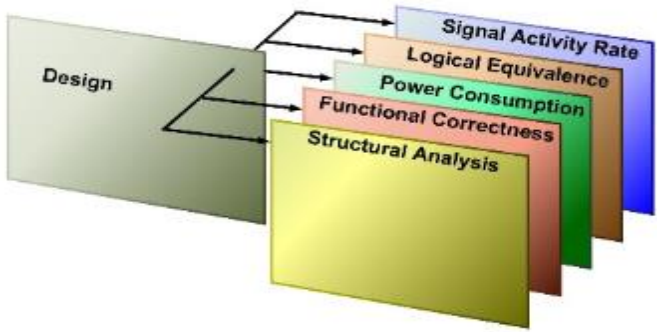

*Figure 7: Breaking down the design into subcategories*

The integrity of a design can be defined as the amount of deviation observed in a one-to-one mapping for each of the subcategories defined above where the actual vs expected is compared to the reference specifications, simulation, and testing results. In Kimura's [35] the subcategories are analyzed individually and normalized, then accumulated together, for a single set of metrics. The DI trust metric considers the design integrity plus the reference design quality. The reference design quality is the set of design artifacts that are obtained. These can range from fully synthesized behavioral models, being the highest reference confidence to the data sheet, being the lowest. Figure 7 shows the trust metric ranges within 0 to .19 being the lowest range and with 1 the highest. Equation 9 is used to calculate the Trust Measure.

$$TM = DI * R \quad eq\ (6)$$

where DI is the Design Integrity and R is the Reference Design Quality.

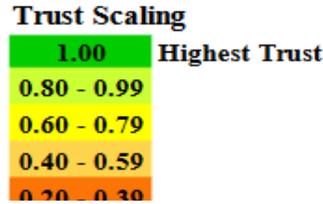

*Figure 6: Trust scaling for hardware components*

**What does it do well** – From the design validation point of view these subcategories represent the areas that are prone to the injection type of attacks and forensic analysis for counterfeits. Therefore, having a trust metric based on hardware design properties provides a degree of assurance that an IC has been validated between the expected design/simulation and the as built or actual. As shown in Table 3 [35] where the test cases where each of the five subcategories are quantified, the Design Integrity metric is calculated and the Figure of Deviation (FOD) is calculated. The top and bottom boxes are inductions that no malice hardware injected faults were detected, the orange one being all signal bits being inverted on the output and the two yellow ones being a counter added to trigger a XOR operation. The other yellow one being an add function with a bit-wise inverter. These were determined by having a test fixture where the expected results were compared to the actual ones.

*Table 3: Test cases for the Design Integrity Analysis*

| Test Article | $SR_{integrity}$ | $P_{integrity}$ | $LE_{integrity}$ | $S_{integrity}$ | $F_{integrity}$ | Design Integrity | FOD | Description of Error Insertion |
|---|---|---|---|---|---|---|---|---|
| TA1 | 1.00 | 1.00 | 1.00 | 1.00 | 1.00 | 5.00 | 0.00% | No malicious content added. |
| TA2 | 0.67 | 0.98 | 0.13 | 0.56 | 0.99 | 3.33 | 33.40% | Counter added to trigger XOR operation on output |
| TA3 | 1.00 | 0.99 | 0.00 | 0.97 | 0.87 | 3.83 | 23.40% | AND function with bit-wise invert |
| TA4 | 1.00 | 0.99 | 0.00 | 1.00 | 0.00 | 2.99 | 40.20% | All signal bits inverted at output |
| TA5 | 1.00 | 1.00 | 1.00 | 1.00 | 1.00 | 5.00 | 0.00% | Stray signal wire added |

This furthers the need to have the foundry be represented in the supply chain metrics that cover the other types of attacks that were mentioned above. The focus of IC design and accelerators like FPGAs are most prominent in autonomous robotic systems.

The use of trust metrics for design integrity is vital for the holistic security model.

**Shortcomings -** While the set of values shown in Figure 7 represents a Trust Metric for hardware design quality, it only covers the logic design part of the process – it does not address the foundry or the design flaw costs that were also discussed in Kimura[35]. As stated, before the IC provider has all the knowledge and artifacts from the development process and this methodology can be applied in house, the consumer of the IC will need to be more resourceful in obtaining knowledge about the design.

In general, relatively few entities are designing the actual components, and more are using COTS devices. Thus, the importance of obtaining components from sources with integrity increases. In the example of the IMU, a complete board is provided by Sparkfun, that may contain components of unknown provenance or a potential risk for compromise in the signal paths. Another potential danger in using a quick to get, cheap and off the shelf component, is that it may work in the prototype phase as desired, but using the prototype component in a production phase will increase the risk of a flaw, like a Sparkfun board. This causes reliability issues and may expose inherent security flaws. As part of the hardware component metrics the need to incorporate the loss and reward values must also be factored in an overall trust model. These two values are described in more detail in the software section. The trust scaling does provide a good set of metrics for the hardware components, because it focuses on the design elements.

V. SOFTWARE LEVEL TRUST EVALUATION

To identify methodologies used for evaluating trust at the software level, we searched for papers focused on BNs, trustworthiness, and security. From these parameters the results are categorized into two groupings standards and the usage of the Common Vulnerability Scoring System (CVSS). We highlight some of the software search results that are attractive to incorporate into our holistic security model.

Software in a robotic system extends from firmware to kernel/Operating System (OS) into middleware and application layers. Software can also be embedded into controllers or using device driver logic within the OS to enable hardware components and can be found embedded in sensors (such as the MPU-9250 mentioned above) or actuators. Software need not only be logically correct but may also need to support temporal constraints. This means that the OS supports a preemptive scheduling on time-bounded tasks, so that the task is completed within a time constraint. The time bound processing is called real-time and this term can be further broken down into hard or soft, where in the case of hard real-time, there is no tolerance for a delayed response and in the case of soft, there are tolerances. Software has far reaching effects at different levels of the stack; therefore, security for it must be taken into consideration.

Software metrics provide a quantitative means to assess or measure the efficiencies in the software development lifecycle, where requirements, design, implementation, testing, and documentation are the different phases. We focus on the implementation since code is an output of this phase. The remaining phases will be considered as part of the supplier chain. The reason why software security metrics are important at the implementation phase is because the code exposes the underlying vulnerabilities. Metrics are used to assess the software assurance against vulnerabilities/weakness from an attack and finding a set of trust metrics is another component to a holistic security architecture.

As an example, consider that the software running on any given robot today is likely a combination of open source and COTS code modules obtained from a variety of public and private sources. For example, OpenSSL is a well-established cryptographic library that provides a number of data protecting functions that are embedded within many operating systems. If it works as expected in a system, we tend to think it is ok, providing the level of security the industry associates with OpenSSL. However, an important assumption underlying the security of communications using OpenSSL is an assurance of the physical security of trusted nodes. For example, the authors have already demonstrated a spy process that can capture encrypted data, as well as the necessary decryption keys, and send them to a remote service without the user knowing their system has been compromised. This demonstrates that even a well-known library, like OpenSSL, can be compromised if the physical hardware is compromised. Since, in an autonomous robotic system, nodes may be captured, an assumption of physical security may be invalid[36].

In our research results we came across a number of standards that define metrics for vulnerabilities, whether in misuse, misconfiguration or in a weakness of implementation. These are described as follows Common Vulnerability Scoring System (CVSS), Common Configuration Scoring System (CCSS), Common Misuse Scoring System (CMSS), and Common Weakness Scoring System (CWSS). Since these are derived from the CVSS standard we will discuss these as a group and will compare them below.

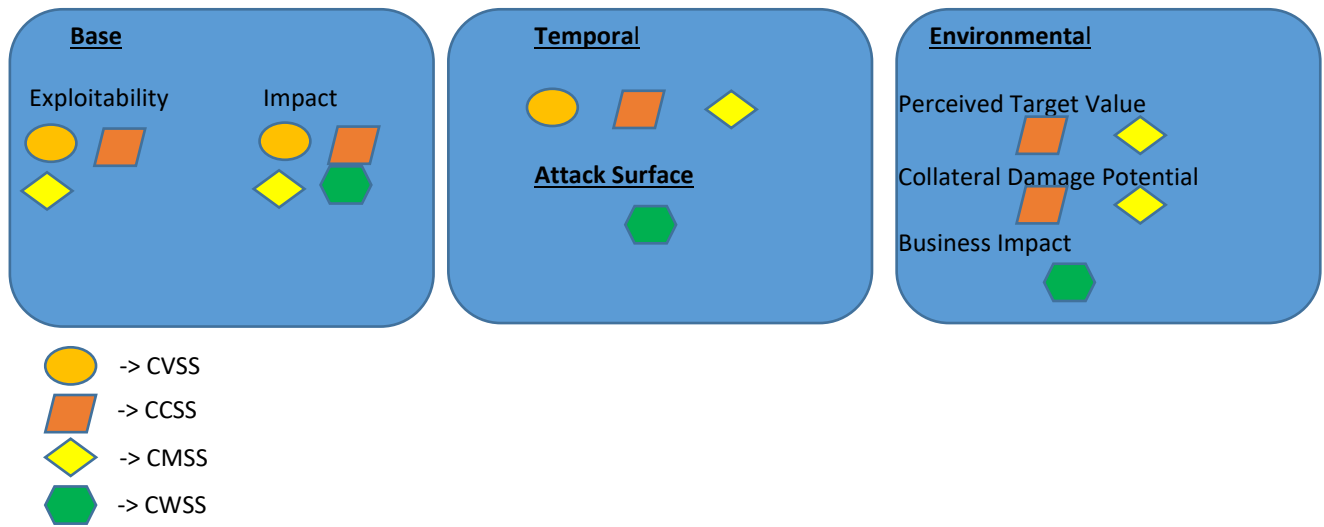

*Figure 8: Comparison of CVSS, CCSS, CMSS and CWSS scoring specifications*

**What is it** - We first start with CVSS, which is the main standard related to the National Vulnerability Database (NVD) that is a repository which provides Common Vulnerabilities and Exposure (CVE) entries. The Common Vulnerability Scoring System is a standard for assessing known vulnerabilities [37]. The next standard CCSS [38] is derived from CVSS but deals with the misconfiguration class of vulnerabilities. Vulnerabilities from misconfiguration can be described as a software feature that enables changing settings. For example, changing access control settings on a directory where Read/Write permission is given to other/everyone via file manager can leave sensitive data exposed, if not used properly. This type of security exposure is deemed a configuration vulnerability. Like CCSS, CMSS [39] is derived from CVSS but deals with the misuse class of vulnerabilities. A misuse vulnerability is built into the software as a feature where the intended feature can expose a security vulnerability. For example, a web link can lead to a malicious site in a messaging application by clicking on the link. The software design feature was to allow a user to follow a link, but not protect against potential exploits. Our final standard, called CWSS is also derived from CVSS but deals with the weakness class of vulnerabilities. CWSS is a part of the Common Weakness Enumeration (CWE) project, co-sponsored by the Software Assurance program in the office of Cybersecurity and Communications of the U.S. Department of Homeland Security (DHS) [40]. An example of a weakness is given by CWE-290 (authentication bypass spoofing), which is caused by an improperly implemented authentication scheme. A spoofing weakness can be caused by only checking the IP address of a client that can be spoofed by the attacker and not using a DNS lookup as the source. A comparison between each of the outlined specifications is that CVSS is based on a vulnerability that is known and CWSS can be used as an assessment metric for software assurance.

**How does it work** – Basically each standard has three groups called Base, Temporal or Attack Surface which is unique to CWSS and Environmental. The Base Group has two metric subgroups; *exploitability* defines the complexity to achieve the exploit and the *impact* defines the result of the exploit. The Temporal group defines the time dependent attributes for the exploit in CVSS, CCSS and CMSS. In CWSS instead of having the Base/Exploitability and Temporal categories some elements are placed into an Attack Surface group. The Environment group defines the surrounding specific attributes for the exploit. In the case of CCSS and CMSS, both Perceived Target Value and Collateral Damage Potential are defined whereas in CWSS a business impact is specified in the Environmental Group. Figure 8 shows the comparison of the four different types of scoring specifications overlaid on to each other.

Each standard has a set of equations for calculating a score and an example of a known vulnerability using the CVSS scoring. As discussed earlier the National Vulnerability Database (NVD) is a repository which provides CVSS base metrics on Common Vulnerabilities and Exposure (CVE) entries. Once a vulnerability is discovered, it is assigned a unique CVE Identifier. In this example, the CVE-2019-15786 Detail ROBOTIS Dynamixel SDK through 3.7.11 has a buffer overflow via a large rxpacket is given with the following parameters.

Using the following vector /AV:N/AC:L/PR:N/UI:N/S:U/C:H/I:H/A:H, a base score is calculated with a 9.8 result being critical. This vector is broken down into the following structure where:

- Attack Vector = Network
- Attack Complexity = Low, Privileges Required = None
- User Interaction = None
- Scope = Unchanged
- Confidentially = High
- Integrity = High
- Availability = H

A calculator can be used to obtain the same value score with these parameters. The specification, examples and calculator are located at the First website [37], the calculator breaks down the three groups into a set of values where metrics are assigned and an overall score is the result. The scores rank from 1 to 10, with 10 being a high severity. For CVSS, the base score is only used in the CVE and NDS entries.

**What does it do well** – CVSS, provides a set of metrics for known vulnerabilities and is used in many types of analysis tools we found in our search. The base impact set of metrics are common values that were used in most of our research results. As for CCSS and CMSS, they provide a set of metrics for Collateral Damage Potential and the Perceived Target Value. These two metrics represent the damage caused by an attack and the reward. These values provide insight into the assessment of an attacker's goals and should be included as part of the trust metric for the different layers of an autonomous robotic system. Finally, CWSS provides a set of metrics for potential weakness in software and hardware that can be used for software assurance. In fact, a number of software tools are using CWE metrics for static and dynamic analysis and the NVD is also using this metric in conjunction with CVEs. CWSS provides a set of metrics that can be used in the software layers for the holistic security metrics.

**Shortcomings** – While CVSS and CWSS are targeted toward conventional known software vulnerabilities, autonomous robotic systems are a new topic for this standard. However, limited vulnerabilities have been found using this technique. As for CCSS and CMSS, while these contributing type vulnerabilities can be found in the CWSS and CVSS as the two complementary specifications, CCSS and MCSS maybe redundant.

From the number of categories and sub-categories from each of the scoring specifications it seems that impact, collateral damage, and perceived target value are the three most important attributes for calculating a trust metric. The other categories and sub-categories are attributes related to supporting/describing the exploit as to when and where.

From the CWSS specification, technical impact scoring provides a better set of definitions with their corresponding coefficients that are used in the equations to derive the category numerical value for assessment (Critical, High, Medium, low and None (1, .9, .6, .3 and 0)[40].

From CCSS and CMSS the collateral damage provides coefficient metrics that are: none, 1, low: 1.25, low-medium: 1.5, medium-high:1.75 and high: 2 [38] [39].

From CCSS and CMSS the perceived target value metrics are: low: 0.8, medium: 1.0, and high: 1.2 [38] [39].

Several articles were found that used attack graphs and utilized CVSS metrics for determining security flaw vulnerabilities. These articles picked a small portion of the system or looked at a network configuration; that was software centric. Current approaches of combining CVSS scores have at least two limitations: they lack support for dependency relationships between vulnerabilities, meaning that they may be ignored or modeled in an illogical way. The other limitation is that they focus on successful attack probabilities only (Cheng, et al) [41]. Frigault et al [42] and Xie et al [43], presented two techniques where the CVSS scores are used as metric values in a BN. The CVSS scores and BN usage model are similar to the technique from Shetty as discussed earlier. Xie et al, extends the usage model by incorporating noisy gates to assist in uncertainty in real-time security analysis. Each node in a BN requires a distribution which is conditioned on its parents. To model the child node's effects from each independent parent node, a combined influence can be achieved by using logic gates (e.g. AND, OR and etc). The term noisy reflects the fact that the logic gate combination is probabilistic and not deterministic.

Frigault et al; presents a technique to measure the overall network security by combining CVSS, attack graph and BN. This technique is further extended into a Dynamic BN where temporal events are represented in the model.

**What is it** – An overall metric is developed for network security using a set of CVSS base scores that are converted into probabilities which are assigned to the paths in an attack graph for an overall assessment value. Both the attack graph properties and probabilities are used in a BN and are extended into a DBN for time-based events.

**How does it work** – – An annotated attack graph is first generated and then a BN is derived from the data in the attack graph. The nodes of the BN represent exploits and conditions derived from the attack graph. Each node represents a probability derived from the CVSS score. Conditional Probability Tables (CPT) represent the causal relationships between exploits and conditions. A BN inference model can reason about if an attacker can reach their goal by any condition. A DBN was also presented that utilized the temporal scores in CVSS for the vulnerability, which showed the time-based activities in exploit maturity, remediation level, and report confidence [42].

**What does it do well** – This approach combines the CVSS (for known attack vectors and paths taken) metrics into probabilities, so that CPT are constructed and utilized in the BN, then extended into a DBN. The use of causal inference is a good method to understand relationships and potential uncertainty.

**Shortcomings** – The scoring mechanism is based on a known attack graph and CVSS score metric but does not consider the attacker's experience/knowledge about the environment. Using CVSS as the only metric is one method of achieving the construction of the BN. This is limited to one data point that states an existing exploit is already found in the NVD and new ones are unaccounted. Since robotic systems is a new domain, these are limited.

Xie et al, presents a technique that extends the CVSS scores and BN usage model by incorporating noisy gates to assist in uncertainty in real-time security analysis. Real-time security analysis in this context is related to observations from an intrusion detection system Intrusion Detection System (IDS) sensor providing false or negative readings or a file system integrity checker such as Tripwire that alerts to a file being changed. The term uncertainty is related to an attack being successful, the uncertainty of an attacker's path choice, and/or the uncertainty from imperfect IDS sensors.

**What is it** - The CVSS scoring metrics allow security analysis tools to define potential exploits in the pre-deployment phase, but how does one account for uncertainties? A BN model is presented that separates three uncertainties in real-time security analysis: the uncertainty on attack success, the uncertainty of attacker choice, and the uncertainty from imperfect IDS sensors.

**How does it work** – Using CVSS metric scores (Base and exploit in the temporal category), the CPT are structured from these values. An attack from an attacker is defined as the physical path (attacks can only occur by following network connectivity and reachability; this is the physical limit for attack) and the attack structure (attacks can only happen by exploiting some vulnerability, with pre-conditions enabling the attacks and post-conditions as the consequence (effect)). A tripwire node is created to sense the detection of an attack, for example from an IDS sensor. The use of the Noisy AND/OR logic are utilized for the conditional events that the attacker must take in order to achieve the vulnerability or goal [43].

**What does it do well** – The use of tripwire and Noisy AND/OR logic is used for the conditional events. The combination of these two logics enables switching for real-time events/analysis. These two logics can be extended into the autonomous robotic layers of the BN for triggering on time events.

**Shortcomings** - Real-time security analysis is a far more imprecise process than deterministic reasoning. We do not know the attacker's choices, thus there is the uncertainty from unknown attacker behaviors. Cyber-attacks are not always guaranteed to succeed, thus there is the uncertainty from the imperfect nature of exploits. The defender's observations on potential attack activities are limited, and as a result we have the uncertainty from false positives and false negatives of IDS sensors[43].

Our search results showed that these techniques provide a basis for the assignment of scores and in aggregate they provide coverage of many things important to score. They miss a few things, that can be easily added including the need to incorporate loss and reward values into their security assessments. The discussion of asset value and loss were represented in Cheng et al[44] with regard to security measurements for situation aware cyberspace. We view that loss and reward should be part of the risk calculations to provide a better qualitative assessment. By capturing the coefficients for impact, cost of loss and reward value, these will be utilized in a trust model that represents the software components related issues. For those things that need to be added, there is not enough experience with robotic systems to form a basis for assignment of scores, but the methodology is helpful.

## VI. Cognitive/AI Level Trust Evaluation

An autonomous robotic system may have one or more AI algorithms running on its system for it to operate within an environment that has many uncertainties. Often, the robot must learn and adapt to the stimuli from the environment. For example, having an autonomous robot in the household as a caregiver aide or assistant. An adversarial attack can alter the robot's behavior by mischaracterizing objects like medicine or producing wrong monitoring results for the patient. This is only a small window of the possibilities of scenarios where an AI is being deployed and having assurance about the implementation is needed, especially in critical and safety applications.

To identify methodologies used for evaluating trust at the Cognitive or AI level, we searched for papers focused on cognitive, trust metrics, BNs, and AI robustness. From these parameters the results were categorized into three groupings: standards [45], [46], techniques for detecting adversarial attacks [47]-[48] and accuracy vs adversarial training. We highlight some of the Cognitive/AI search results that are attractive to incorporate into our holistic security model. AI certainly is not new; it has been around since the mid-1950s when John McCarthy used the term Artificial Intelligence and invented the LISP programming language. The path of AI has taken different twists along the way with expert rule base systems as one form, to neural networks that mimic the brain, and currently the combination of these two called symbolic AI. In order to handle complex data for AI models to be useful the community is leveraging Convolution Neural Networks (CNNs). CNNs use a form of deep learning that is targeted toward image and natural language domains. Reinforcement learning is another form of Deep Neural Networks (DNN)s targeted toward path planning, while perception and motion are often associated with autonomous mobile platforms. The new focus of research is AI adversarial attacks, the classification of data being poisoned, evasion attacks and black box attacks to name a few. For every attack, a remedy may arise to counter, but this takes time and there needs to be a method to identify these types of attacks. In the case of autonomous mobile robots, an AI/learning layer opens up new types of attack strategies that more conventional attacks do not.

AI logic often comes from open or COTS sources and determining the robustness to adversarial attacks needs to be considered for the system trust model. If these AI implementations have not been subjected to some adversarial data training, the lack of testing with different types of data sets will lead to an untrustworthy implementation. The need for a robustness metric is required, so that certain implementations can be used in high assurance scenarios. Adversarial attacks have been discussed in the literature related to poison (tainting the training data), white/black/grey box (using physical and alternative training models), and evasion (misclassification by spoofing). A large portion of research has been geared toward image-based data sets; however, testing should be expanded into other domains. Since AI robustness and adversarial attacks are new, our research has come up with a mixed bag of results that are grouped into possible solutions, the first group is adding adversarial data to the training cycle, the second group is using a value or utility function to assist in a distance measurement functions and the third group is using linear bounds as constraints. We first describe the types of attacks to provide an overview for why AI robustness metrics are needed, since this is an evolving field.

An example of a poison type of attack is in spam filters, where emails are filtered for characteristics of being spam. What looks like a normal email is used to train the spam filter's algorithm to include the extra words to be recognized in its knowledge base. This process then causes the spam filter to flip what would normally be seen as a good email into a bad one.

White, black, and grey box attacks are related to an adversary having different levels of knowledge about the system. In the case of white box, the adversary would have full disclosure about the algorithm including the weights and all data used during training. An example of a white box attack is in object recognition systems like handwritten symbols in an image where a 1 can be recognized to be a 7 with small modification of the image. A black box case is where the adversary has no knowledge of the algorithm or data. An example of a black box attack is subjecting the model being attacked to different inputs and acquiring the outputs, the adversarial knowledge that is obtained can be transferred into a substitute model for further exploitation. In the case of grey box attacks, the perturbation of a pixel can cause misclassification of an image like in the stop sign example discussed below.

A simple case of evasion attacks is to manipulate the test data so slightly that it does not change the classification boundary and in the spam filter case the email is obfuscated so that it passes detection. These are only a few examples of adversarial attacks.

To assist in adversarial defenses, DARPA has a new program called Guaranteeing AI Robustness against Deception (GARD) that was announced in early 2019. GARD goals are to come up with metrics that measure adversarial perturbation and the capability to defend against several attacks. Figure 9 shows a cycle of attacks and resulting defenses but points out that these defenses do not generalize to new attacks. For example, each attack corresponds to a single defense, represented in Figure 9, where the bottom left-hand side starts with an attack and right-hand side is the related defense [49].

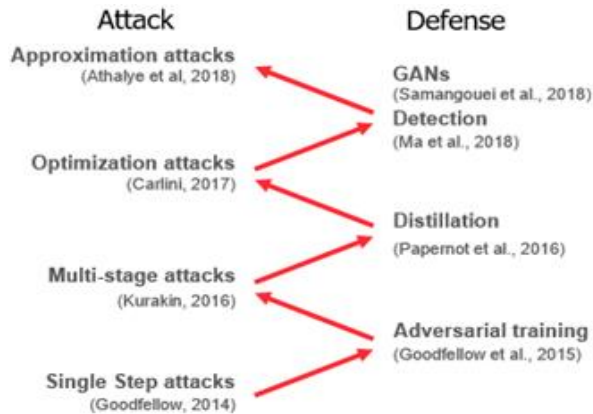

*Figure 9: Single Attack vs Defense for Machine Learning*

Machine learning classifiers are tuned to make decisions from training data, like an algorithm being able to recognize a stop sign from an image data set. There are different types of classifiers that span predictive, binary, nearest neighbor, support vector machine and deep learning to name a few. These classifiers create boundary lines from the data they receive that helps recognize the object that it was trained for. Adversarial perturbation changes the boundary line, and this leads to mischaracterization of the object. This is seen in Figure 10 where the tight boundary that distinguishes between the data types are shown [49].

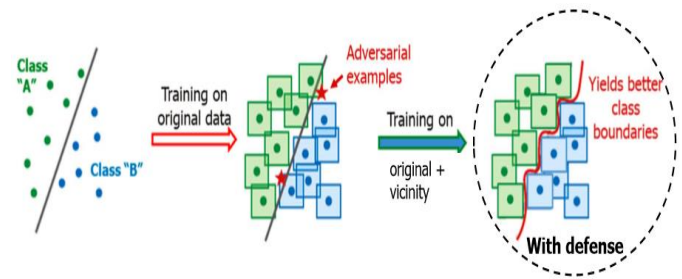

*Figure 10: Creating defense boundaries*

An adversarial example is recognizing a stop sign, but when the sign is slightly changed a mischaracterization occurs and, in some cases, this can cause grave danger. In Figure 11 a normal stop sign is seen on the left, but, on the right, slightly changing the text on the sign causes a resulting mischaracterization of the sign as indicating a 45-mph speed limit.[50].

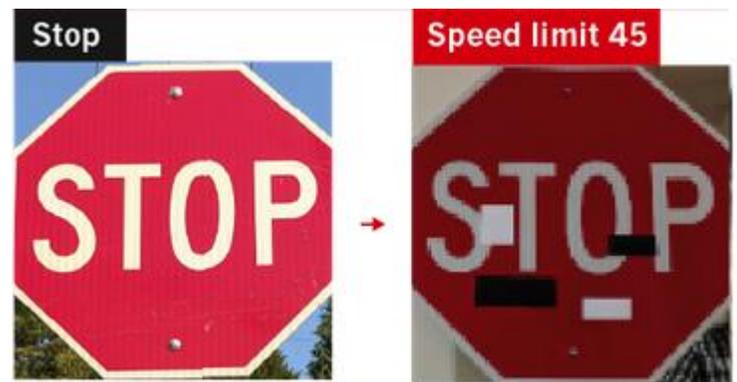

*Figure 11: Altered text on stop sign*

Another adversarial example is targeting neural network policies in reinforcement leaning called adversarial policy. The policy is a set of action rules the agent can take as a function of state and environment. This policy causes the multi-agent environment to generate seemingly random and uncoordinated behavior. This behavior can be seen in Figure 12 where the victim (in blue) is against a normal opponent on the top and an adversarial opponent on the bottom[51].

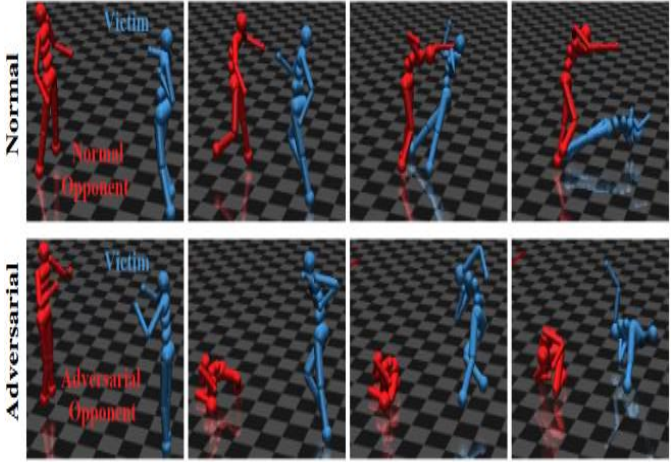

*Figure 12: Adversarial policy in deep reinforcement learning*

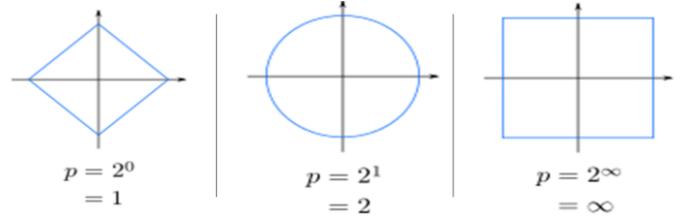

*Figure 13: Minkowski's distance metric for p-norm*

Based on the above we take a deeper look at these adversarial examples. In the stop sign example shown in Figure 11, the left-hand side is recognized as x and classified originally as target t = arg-max F(x). The right-hand side is a new desired target where t′ not equal to t. This is called x′ a targeted adversarial example if arg-max F(x′) = t′ and x′ is close to x given a distance metric [52]. Another type of adversarial example is called the Fast Gradient Method (FGM), which is a one-step algorithm that takes a single step in the direction of the gradient. $x' = FGM(x) = x + \epsilon\, sign(\nabla x\, J_{(\theta,x,y)})$, where θ is the parameters of a model, x is the input to the model, y is the targets associated with x, J(θ,x,y) is the cost used to train the neural network and ∈ controls the step size taken [53]. The cost function can be linearized around the current value of θ and obtain an optimal max-norm constrained perturbation. To determine where boundary lines are placed, a number of techniques are used for distance metrics in order to differentiate between *x* and *x'*, original and perturbed data. We highlight three techniques, first is Minkowski's; general formula, second is Lipchitz continuity and the third is an activation function for finding or determining distance functions.

The minimum distance of a misclassified nearby adversarial example to x is the minimum adversarial distortion required to alter the target model's prediction, which is referred to as the lower bound. A certified boundary guarantees the region around x that the classifier decision cannot be influenced from all types of perturbations in that region. In other words, the robustness is being able to detect perturbation as close to *x* as possible and, in some cases, this is an approximation or an exact guarantee to determining the lower boundary point. In order to evaluate the distance, sometimes called distortion or error between x' and x, the generalized Minkowski's formula is used to calculate the distance metric within p-norm space. For p-norm when p=1, this is a Manhattan distance, when p=2 is the Euclidean distance, and p=∞ a Chebyshev distance. This is shown in Figure 13 below.

$$D(X,Y) = \left(\sum_{i=1}^{n} |x_i - y_i|^p\right)^{\frac{1}{p}}$$

Another technique that is used is called the Lipchitz continuity for determining distortion between measured space. The equation of $|f(X_1) - f(X_2)| \leq K |X_1 - X_2|$, where two metric spaces are given and if there exists a real constant where $K \geq 0$, for all $x_1$ and $x_2$, K is referred to the Lipchitz constant.

The activation function technique is when a neuron or node fires in a Neural Network (NN) that allows the input data to pass (directly or transformed) to the output stage depending on the layer depth. Activation functions can be linear or non-linear, where the latter can handle more complex data types. Rectified Linear Unit (ReLU), hyperbolic tangent, sigmoid and arctan are several types of activation functions. In relations to digital circuits, this is the rising edge that turns a gate on or off, but with different ranges and signal characteristics. NN have building blocks that are used to transform data from input to output nodes or to different layers. A Pooling block reduces the number of parameters and computation in the network by reducing the spatial size of the data represented. A Residual block feeds the next layer but also supports a jump in layers from 2 to 3 hops away and a Batch normalization blocks provides a standardized technique for inputs to the next layer, so that a reduction in training cycles and learning process can be achieved. A newer technique is used by applying two linear bounds on each activation function where the output of each layer is constrained by these two terms. We present some of our search results that utilize some of these techniques to determine an approach to finding a robustness metric in three different groupings.

One group tried to dwarf adversarial attacks by increasing the accuracy of the model using the p norms, so that both adversarial and training data combined would be used to train the model. But the findings from Tsipras et al [54], Nakkiran [55], and Su et al [56] pointed out the need to differentiate between the two types of training (classifier and adversarial) and not make the classifier so tuned to adversarial data, because the model's overall accuracy decreases.

In another group of findings, the authors Weng et al [57] derived their work from Hein and Andriushchenko's [58] using Extreme Value vs Mean Value Theory and Lipschitz continuity for determining lower and upper bound approximation for adversarial detection. Weng et al, presents the Cross Lipschitz Extreme Value for nEtwork Robustness (CLEVER) for determining a robustness metric.

**What is it** – CLEVER is a technique for determining the lower bound of a neural network that achieves a robustness metric against adversarial attacks. In other words, it determines the minimal distortion level to achieve an adversarial attack from the original data.

**How does it work** – This is the intuitive explanation of how finding the lower bound is performed using Lipchitz continuity in the paper. In Figure 14 the function value g(x) = fc(x) - fj(x) near point $x_0$ is inside a double cone formed by two lines passing ($x_0$; g($x_0$)) and with slopes equal to ± $L_q$, where $L_q$ is the (local) Lipschitz constant of g(x) near $x_0$. This means that the function value of g(x) around $x_0$, for example g($x_0$ + δ) is bounded by g($x_0$) given δ (adversarial perturbation) and $L_q$ (Lipschitz constant). When g($x_0$ + δ) is decreased to 0, an adversarial example is found and the minimal change of δ is the distortion difference between adversarial example and original data) is given by g($x_0$)/ $L_q$ [57]. The cross Lipschitz constant is the cross terms fc(x) - fj(x) of the function.

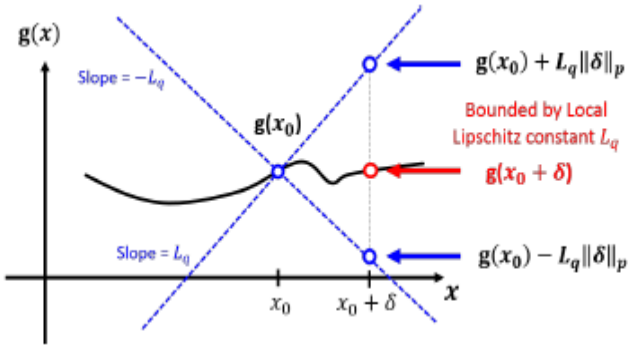

*Figure 14: Lower bound explanation*

There are two algorithms that are defined, the first algorithm is for targeted attacks and the second is the same except for removing the target class that suites the non-targeted attacks. The goal of the targeted attack is to make the model misclassify by predicting the adversarial example, as the intended target class instead of the true class. The untargeted attack does not have a target class, but instead it tries to make the target model misclassify by predicting the adversarial example as a class, rather than the original class. Targeted was defined as three different classes called random target, least likely, and top-2. The random target class is defined as randomly selecting a target. The least likely class is the lowest probability when predicting the original example. The top-2 class is the highest probability except for the true class, which is usually the easiest target to attack. Basically, sample a set of points within a circle boundary using Lp=2 as in step 4, then find the max value and store that in S for each group of points or batch. In step 9, a maximum likelihood estimation is performed on the reverse of the Weibull distribution parameters (a Cumulative Distribution Function (CDF) has a finite right endpoint (denoted as $a_w$). The right endpoint reveals the upper limit of the distribution, known as the extreme value. This equation is the inverse Weibull distribution, where G(y) is the CDF of max y's or limit distribution and $a_w$, $b_w$ and $c_w$ are the location, scale and shape parameters, respectively.

$$G(y) = \exp\left\{-\left(\frac{a_w-y}{b_w}\right)^{c_w}\right\}$$, where $y < a_w$ and when $y > a_w$, it becomes 1.

The extreme value is exactly the unknown local cross Lipschitz constant [57].

**Algorithm 1:** CLEVER-t, compute CLEVER score for targeted attack
**Input:** a K-class classifier $f(x)$, data example $x_0$ with predicted class $c$, target class $j$, batch size $N_b$, number of samples per batch $N_s$, perturbation norm $p$, maximum perturbation $R$
**Result:** CLEVER Score $\mu \in \mathbb{R}_+$ for target class $j$
1  $S \leftarrow \{\emptyset\}, g(x) \leftarrow f_c(x) - f_j(x), q \leftarrow \frac{p}{p-1}$.
2  **for** $i \leftarrow 1$ **to** $N_b$ **do**
3  $\quad$ **for** $k \leftarrow 1$ **to** $N_s$ **do**
4  $\quad\quad$ randomly select a point $x^{(i,k)} \in B_p(x_0, R)$
5  $\quad\quad$ compute $b_{ik} \leftarrow \|\nabla g(x^{(i,k)})\|_q$ via back propagation
6  $\quad$ **end**
7  $\quad$ $S \leftarrow S \cup \{\max_k\{b_{ik}\}\}$
8  **end**
9  $\hat{a}_W \leftarrow$ MLE of location parameter of reverse Weibull distribution on $S$
10 $\mu \leftarrow \min(\frac{g(x_0)}{\hat{a}}, R)$

**Algorithm 2:** CLEVER-u, compute CLEVER score for un-targeted attack
**Input:** Same as Algorithm 1, but without a target class $j$
**Result:** CLEVER score $\nu \in \mathbb{R}_+$ for un-targeted attack
1 **for** $j \leftarrow 1$ **to** $K, j \neq c$ **do**
2 $\quad \mu_j \leftarrow$ CLEVER-t($f, x_0, c, j, N_b, N_s, p, R$)
3 **end**
4 $\nu \leftarrow \min_j\{\mu_j\}$

To illustrate this further Figure 15 [57] is a two-dimensional space with three hyperplanes and the two balls (p norm) are bounded by a radius. The three hyperplanes $w_ix+b_i = 0$ divide the space into seven regions (with different colors). The red dash line encloses the ball B2(x0;R1) and the blue dash line encloses a larger ball B2(x0;R2). A sampling is done within the two ball spheres as to finding the max y's.

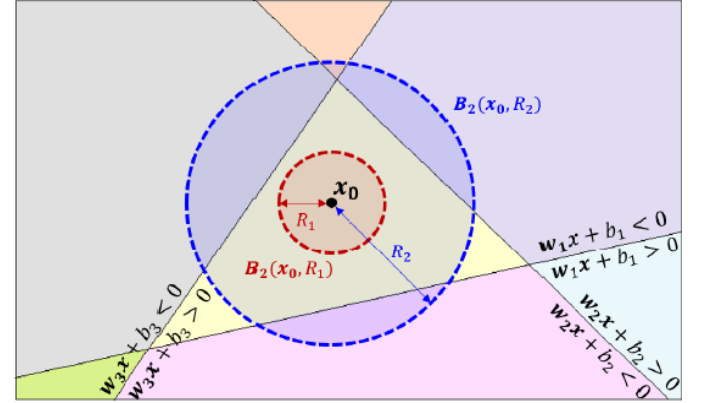

*Figure 15: Illustration of the algorithm*

**What does it do well** – CLEVER provides a means to approximate a formal guarantee on AI robustness by finding the lower bound on an implementation for detecting adversarial perturbation. This is useful in understanding the security limits of AI algorithms that run on autonomous robotic systems. CLEVER is attack-agnostic, works for large neural networks, it is computationally feasible.

**Shortcomings** – CLEVER was validated using image data and tested against a few other adversarial examples. This small sampling of tests needs to be expanded into a larger test space for this to be considered a generalized solution. However, it does provide a step in the right direction to a qualitative metric for AI robustness against adversarial attacks.

A third group leveraged the work of CLEVER as an approximation for defining a lower bound on adversarial perturbation detection, where CROWN (another detection technique) extends to create an exact certification for general activation functions in DNN. This same research group proposes the CNN-Cert, a certifying framework to accommodate general DNNs [59]. This technique focuses on using two linear bounds constrained on activation functions for each output layer.

**What is it** – Boopathy et al, presents a framework for certifying robustness in neural networks with guarantees on adversarial attack detection. This framework is called CNN-Cert for Convolution Neutral Networks, but the authors state that it can support a number of other architectures including max-pooling layers, batch normalization, residual blocks as well as general activation functions [59].

**How does it work** – Let f(x) be a neural network classifier function and $x_0$ be an input data point. They use σ(·) to denote the coordinate-wise activation function in the neural networks. Some popular choices of σ include ReLU:σ(y) =max(y,0), hyperbolic tangent: σ(y) = tanh(y), sigmoid: σ(y) = 1/(1+e−y) and arctan: σ(y) = tan−1(y). The symbol ∗ denotes the convolution operation and Φr (x) denotes the output of r-th layer building block, which is a function of an input x. Also, denote Φr−1 as the input of activation layer. The superscripts denote index of layers and subscripts to denote upper bound (U), lower bound(L). The general form of the equation is as follows:

$A_L^0 * x + B_L^0 \leq \Phi r(x) \leq A_U^0 * x + B_U^0$, where $A_U^r, B_U^r, A_L^r, B_L^r$ are constant tensors related to weights and bias as well as the corresponding parameters in the linear bounds of each neuron [59]. This general form is applied to the different residual blocks, pooling layers, and batch normalization layers by deriving the linear upper and lower bounds using the right-hand and left-hand sides of the general equation. CNN-Cert was tested against different image datasets using MMIST and CIFAR (consists of 60000 32x32 color images in 10 classes) using 4 to 7 layers and different filters.

**What does it do well** – CNN-Cert provides a guarantee to finding a certifying robustness region or the minimal distortion detection using linear bounds on outputs. CNN-Cert was tested against different image datasets using MMIST and CIFAR using 4 to 7 layers and different filters with > 11% improvement in performance and compared to CLEVER. Since CNN is a form of DNN, this technique can be applied to different autonomous robotic algorithms to determine the certified region or lower bound.

**Shortcomings** – Similar to the comments above on CLEVER, since this is from the same research group.

From the two different approaches for determining a certified region for detecting adversarial perturbation, one being an approximation and the other being an exact guarantee in AI DNNs, we can derive a rating for AI implementations. By using these lower boundary techniques, a minimum distortion level is established and from this point we can define ranges for rating AI implementations against these known values. To better illustrate this concept the following Figure 16 shows a center region aa equal to the certified region and each subsequent ring are corelated to the rating or strength of the AI implementation using a distance function. Let x be the certified region and y be the AI implementation, we can use the p-norm distance equation to determine the differences for adversarial perturbation detection. We call $R_m$ the AI Robustness Metric for rating the strength of an AI implementation.

$R_m = || x- y||$, where y > 0 and $0 < R_m \leq 1$

As one moves further away from the certified region the rating should decrease within 1 being the lowest

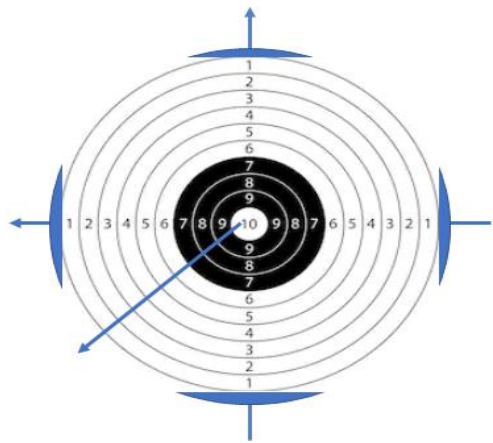

*Figure 16: Robustness distance metric*

The capability to detect the smallest perturbation distortion near *x* (original) provides a higher robustness metric score. In the case that CLEVER score is attack-agnostic, a higher score number indicates that the network is likely to be less vulnerable to adversarial examples; the same applies to CNN_Cert [60]. So, with finding the lower bound, one can start to formulate a way to express a rating system for how AI robustness metrics can be applied to different AI implementations. Since each defense technique has a distance/error from the certified area (boundary) where perturbations can be detected, we can consider these values as trust metrics in a continuous set of ranges between [0,1] as part of the holistic trust model.

Unlike some of the other measures we have seen so far, this is still an area that is less mature. However, in the spirit of the previous discussions, we can consider a system with no "defenses" against attack a more vulnerable system then one that defends itself against multiple types of attack. Therefore, for our purposes we can consider it the same way we are considering other metrics. AI trust metrics and costs will need to be fused together for a set of metrics that will be considered for the holistic model.

Another important element to consider is the training and test data that is being used and how this is being protected for distribution. The supplier of AI components should have a repository of this data, but how is this protected and how is the authenticity of the data checked? The AI component vendor will need to be considered under the supplier trust metrics category as well, since the training and test data will need to be obtained for validation/certification.

## VII. Supply Chain Level Trust Evaluation

A supply chain is defined as set of resources and processes that upon placement of a purchase order begins with the sourcing of raw material and extends through the manufacturing, processing, handling and delivery of goods and related services to the purchaser [61]. Supply chain vendors touch many components of an autonomous robotic system. Fundamentally, these vendors supply all the components used in a system, with individual vendors supplying anything from a single hardware or software component (including AI) to a complete system. Therefore, to ensure the trustworthiness of a system, each of the components or systems must come from entities that are reputable, provide reliable products, support services, and follow best security practices. The system, hardware, software, and AI layers in our holistic security architecture may present security risks from their suppliers. We believe that the supplier is an important entity to incorporate into the holistic security architecture.

A supplier can produce either a whole system or a system of systems as a component within a larger system. A security threat at the system level can be viewed in this example as a drone manufacturer called DJI where backdoor exploits were found that could divert drone data to a remote server or control it. The problem with DJI's backdoors, however, is that hackers or government actors can install malware through these backdoors for cyberspying [62]. This is compounded by the fact that this technology likely has a Remote Access Trojan (RAT) embedded in it [62].

For the hardware level as we discussed in the evaluation of hardware section, security threats can come in the form of malicious insertions where add, delete or modifications of gates can occur at the foundry. For example, a hardware attack technique is called TrojanZero where the researchers were able to insert additional gates into the design without being detected, using standard power analysis at post manufacturing testing phase [63]. They were able to create low probability of triggering a detection with the additional gates as well as enabling backdoor entry. The success of TrojanZero was predicated on using 3$^{rd}$ party testing service for the IC, where the attacker at the foundry had knowledge about the design and testing methodology. A similar hardware attack example can be applied at the Printed Circuit Board (PCB) level where malicious insertion can happen at the raw card manufacturer as discussed by Russ and Gatlin [64].

A software supply chain has several attack vectors that may span from insider malware injection into the code itself, abusing the code signing mechanism, malware injection into the software update mechanism or service, attacking open source directly, and attacks the distribution service like application stores. Some of these examples are CCleaner, SimDisk and ShaowPad where state actors attacked the supply chain and for code signing abuse attacks like ShadowHammer, Naid/MCRat and BlackEnergy 3. Examples of update services being attacked were Flame, CCleaner and Adobe pwdum. Several examples of opensource being affected were RubyGems backdoor, and JavaScript backdoor. A survey was performed that outlined the software supply chain security threats that included the examples mentioned above called " Breaking trust: Shades of crisis across an insecure software supply chain" by Herr et al [65]. As a supplier of software product or service, the software development methodologies must be reviewed with the emphasis on secure assurance engineering as discussed in Section II.

Since the topic AI/Cognitive algorithms for autonomous robotic systems is new, we have to consider the attack vectors at the supply chain. AI is a little different from software since training data, adversarial training data, robustness data, and the models themselves need to be considered as part of the packing or hosting services and delivery of these components. Another consideration is 3$^{rd}$ party components being infected with malware or created potential backdoor exploits that are integrated into the base solution. There is a clear need to implement security controls in the supply chain assessment and formulating a supply chain trust metric will need to account for this feature.

To identify methodologies used for evaluating trust at the supply chain level, we searched for papers focused on supply chain trust metrics, vendor trust metrics, and supply chain risks. From the search parameters the results were categorized into two groupings: standards and different approaches to defining supplier trust metrics [66]. We highlight some of the supply chain search results that are attractive to incorporate into our holistic security model.

One standard, called ISO **28001,** provides a guide for best practices for implementing supply chain security, assessments, and plans [61]. Figure 18 shows a high-level ISO 28000 security management system for a supplier and ISO 28001 provides an eight-step methodology for security risk assessment and development of countermeasures [67]. A description of the security management is first discussed and followed by the eight-step methodology for risk and countermeasures.

- A security management policy includes the organizations charter and goals toward security and the security controls framework. The goals may consist of the overall threat and risk management framework, comply with legislation, regulatory and statutory requirements aligned with the business segments, and clearly document the overall policy/goals and communicate to stakeholders both internal and external including 3$^{rd}$ party.

- The security planning and risk assessment is geared toward if an event occurs what are the steps and process for the crisis. An event can be a physical failure threats and risks (incidental damage, malicious damage or terrorist or criminal action) and an operational threats and risks (security controls, human factors, equipment safety and business impact to operations).

- The Implementation and operations are defined as the infrastructure to support the security controls for the business, create a chain of command, communicate and train personnel and fully documented the policies, objective, and procedures. The operational controls as well as data controls should be implemented with archiving or off-site capability in case of an emergency or security incident. Within the operational and data controls there should be key management controls.
- Checking and corrective action should be taken that affects the security performance measurement and monitoring of the systems, security related failures, controlling the records and audit.
- The organization shall have periodical reviews of the security management systems and implement improvements when necessary.

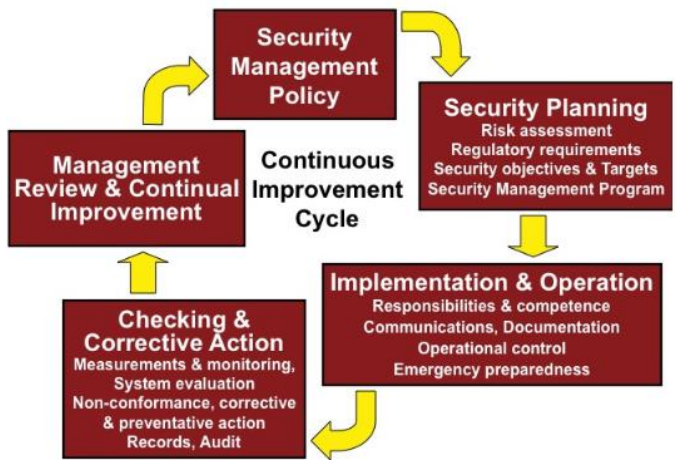

*Figure 17: Components of a security management system using ISO28000*

The eight-step methodology for security risk assessment and development of countermeasures is defined as a vulnerability assessment or analysis and the steps to perform the analysis are described below.

- Step one- is the consider the security threat scenarios, so that each scenario is broken down and analyzed
- Step two – From the analysis determine the consequences and classify them
- Step three – Rate each scenario and associated security incidents with a probability of success
- Step four – Create a security incident scoring for the business
- Step five – From the security scenarios and incident create countermeasures
- Step six – Take results from five and implement them
- Step seven – Evaluate step six
- Step eight – repeat the process

A supplier that does not support a security management system will be exposed to a number of vulnerabilities. Autonomous systems are complex, depending on the functions that they perform. For example a vehicle can have 70 to 100 embedded processors and a plethora of sensors for object detection/navigation and let alone the communications network [68] [69]. Many of the components are COTS and open source software is commonly used. The following packages are the top open source projects used in autonomous vehicles Autoware, Apollo, EB Robinos & EB Robinos Predictor, NVIDIA® DriveWorks and OpenPilot [70]. A new autonomous operating system is open sourced based on ROS 2 called Apex [71]. Therefore, supply chain trust metrics are important, and several factors should be considered when selecting a set of quantitative values for assessing supply chain trust.

Combining the security best practices as mentioned above within the attributes that a supplier might be rated against are shown in Figure 18 [72]. In addition, the following items should also be included in the supplier assessment: the quality of the product/services, reliability, and maintainability of the product/deliverables. Finally, supplier assessment should include the security of the item that also includes the organization security plan, security incidences (including history), any penalties that have been incurred, and financial stability. This combination provides a supplier trust metric that covers the autonomous robotic system layers at a system level or at the individual component level. A Supplier Assessment Management System (SAMS) provides objective measurement criteria for supplier performance in eight major categories.

| Management | Proposal |
|---|---|
| • Responsiveness | • Team Commitment |
| • Program Management | • Proposal Strategy |
| • Risk/Opportunity Management | • Proposal Adequacy and Negotiation |
| • Expert Staffing | |

| Technical | Cost |
|---|---|
| • Product Performance | • Cost |
| • Systems Engineering | • Financial Health |
| • Software Engineering | |
| • Logistics & Sustainment | |
| • Part Materials & Processes | Schedule |
| • Service Level Performance | |

| Mission Assurance & Quality | Supply Chain Management |
|---|---|
| • Quality | |
| • Process Effectiveness | Customer Satisfaction |

*Figure 18: Supplier Assessment Management System (SAMS): 8 Categories of Assessment*

As a result of our research for supplier trust metrics, the SAP score card is presented as an example set of values to be used for metric values.

**What is it** – A description of a supplier assessment score cards are presented by Systems, Applications and Products (SAP) and Northrop Grumman.(NG) [72] that utilizes a framework called the Supplier Assessment Management System (SAMS).
**How does it work** –
The eight categories in the SAMS framework are rated with the following values, using SAP values (blue= 100 to 91, green=90 to 75, yellow=74 to 51 and red=50 to 0). NG Supplier Scoring on each category is performed using these metrics (Unsatisfactory < 2, Marginal 2 to 2.7, Satisfactory 3.75 to 2.76, and Excellent 4 to 3.76).
**What does it do well** – This provides a set of metrics that can be used for the supply chain vendors related to a system level or individual component level of an autonomous robotic system. By using the numbering of the SAP score card and reference labels from NG score card, the result is a combined set of trust metric values that can be used in our model.
**Shortcomings** – While this provides a good set of metrics for the supplier vendor, this will also need to cover the AI components that have been described above. The topic of security will need to be introduced to suppliers that are not familiar with it and the security management system.

We have covered the layers of an autonomous robotic system that included the system, hardware, software, and cognitive layers that will have a supply chain trust metric associated with its components in the holistic security model. By combining the architecture layers and supply chain trust metrics we capture the origins of the components and reveal the risk beforehand vs after system deployment.

## VIII. ASSESSMENT TECHNIQUES

From our search results for each individual evaluation layer of the holistic architecture several assessment techniques have been discussed. These assessment techniques have ranged from pen and paper assessment of items on a checklist to visualization/automating the assessment process using graphic tools in both a static and dynamic manner. For simple systems pen and paper works well Performing the evaluation can be done in a number of ways but falls apart quickly as complexity (and accounting for system interactions) increases. The reason we are talking about assessment techniques is that having metrics is only one part of the puzzle. The next part is evaluating a system based on those metrics. Performing the evaluation can be done in a number of ways, as for visualization/automating there are different techniques that have been applied, including the use of Attack Graphs, Fault Trees, Petri Nets, BNs, and DBNs.

The first assessment technique is the Attack graph for visually representing the asset and target related to the security domain.

**What is it:** Attack Graphs/Trees are conceptual diagrams showing how an asset, or target, might be attacked. These graphs are generated by an analyst/automation that has obtained data from host scanning tools and network diagrams that includes connectivity between hosts. The paths are assessed and ranked from attacker to target using the connectivity connections.

**How does it work:** Attack trees are multi-leveled diagrams consisting of one root, leaves, and children. From the bottom up, child nodes are conditions which must be satisfied (true/false) to make the direct parent node true; when the attack condition at the node is true, the attack is complete. Each node may be satisfied only by its direct child node. An example of an Attack Graph/Tree is shown in Figure 19 [73] [74] which demonstrates how an attacker is able to conduct a series of exploits on Secure SHell (ssh). The ssh is used to encrypt network services like remote command-line, login, and remote command execution. The sshd is the daemon service for SSH and the bof is the buffer overflow on a sequence of host computing devices. The sshd_bof exploit, acquires user privileges (user) on each box. This example shows the sequence of exploits and the preconditions (a set of system properties that must exist for an exploit to be successful and in this case the sshd being executed on the host) that are necessary for a successful attack.

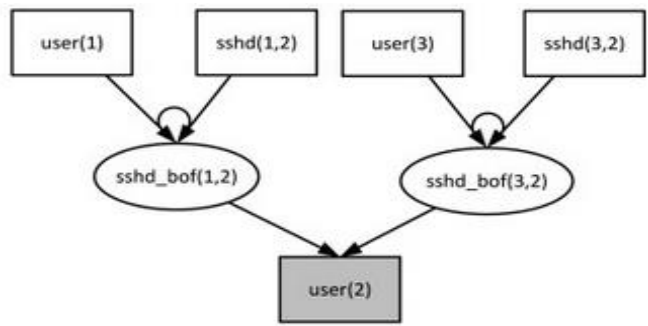

*Figure 19: An example of an Attack Graph*

**What does it do well:** An easy way to represent potential attack vectors is using a tree diagram that can be automatically generated using analysis tools. This will need to be extended for the autonomous robotic system security architecture. This will be considered as part of the NVD data being assessed for the robotic system when CVEs are available.

**Shortcomings:** An attack graph can be exhaustive (all possible outcomes must be known) for analyzing attack vectors. As the number of nodes grows, it becomes difficult both in resources and time to perform the analysis. However, the simplicity of these graphs makes it ideal for security assessment, when isolating the network paths and reduced nodes to analysis.

The next technique is similar to an Attack graph but is called a Fault Tree.

**What is it:** Fault Tree is a top-down, failure analysis in which an undesired state of a system is analyzed using Boolean gates to combine a series of lower-level events [75]. One method of creating a Fault Tree is by following the steps for a Method Of Cut Sets (MOCUS) [76]. MOCUS provides a deterministic result that requires less resources to calculate the top event's probability, thus reducing error and improving performance.

1. Create a table where each row of the table represents a cut set and each column represents a basic event in the cut set.
2. Insert the top event of the Fault Tree in the first column of the first row.
3. Scan through the table, and for each Fault Tree gate:
a. If the gate in an AND gate, then insert each of its input in a new column.
b. If the gate is an OR gate, then insert each of its input in a new row.
4. Repeat step 3 until all the gates in the Fault Tree is explored and the table only contains the basic events.
5. Use Boolean laws to remove all redundancies within the table.

**How does it work:** The Fault Tree Analysis method is mainly used in safety and reliability engineering to understand how systems can fail, to identify the best ways to reduce risk, and to determine/estimate failure event rates or a particular system level (functional) failure. In the case of security, the same example from the Attack graph above is reconfigured for a Fault Tree analysis as shown in Figure 20 [74] [75], where the gates (AND/OR) are used.

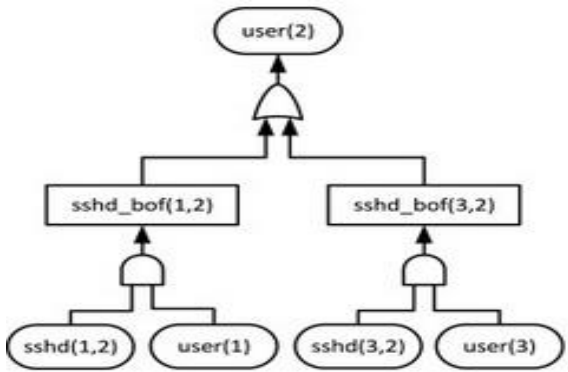

*Figure 21: An example of the fault tree*

**What does it do well:** Fault Trees provide a visual graph of the system that represents relationships between events and their causes. While this is a common model for fault analysis, it can still be used for assessing trust in a system. As seen by the figures, the comparison between an Attack graph and a Fault Tree is similar.

**Shortcomings:** In utilizing Fault Trees, a large tree needs to enumerate all possible sequences of failures in a complex system and has limitations. These limitations are a result of the exhaustive computational resources required to produce an output for all sequences of failure. Classic Fault Tree models that are combinatorial like, Attack and Fault Trees, do not support situations that have complex dependencies at the system or sub system levels. These dependencies are failure characteristics such as functional dependent events and priorities of failure events.

The Petri net technique is different from the first two, because it can represent dependencies transitions with arcs in behavioral control where in the other two cases this was not feasible.

**What is it:** A *Petri net* is a graphical tool for describing the control flow behavior of concurrent processes in systems, which was introduced in 1960's.

**How does it work:** Petri Net is a directed bi-graph and is defined as a 5-tuple N = (P, T, F, W, M0), where P is finite set of places, T is a finite set of transitions, F is a set of arcs, W is a weight function and M0 is an initial marking. Transitions (events that may occur) are represented by bars and places (conditions), which are represented by circles. The directed k-weighted arcs (a measure of the arc multiplicity*)* describe which places are pre- and/or postconditions for which transitions (signified by arrows). The state of a process is modeled by tokens in places and a state is also called a marking, as shown in Figure 21 [77]. Petri nets are concurrent, nondeterministic models, meaning that they can support multiple transitions of events at the same time and the firing of the events can occur at different orders. This example can be shown in Figure 16 where T1 is splits into p2 and p3 paths.

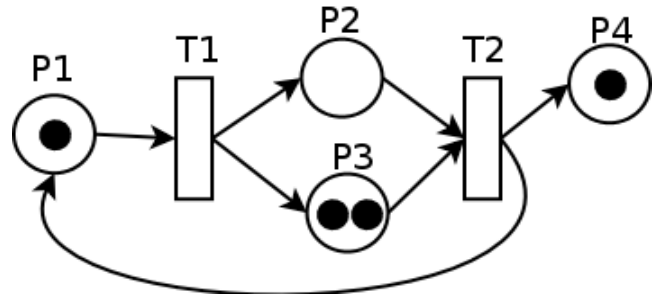

*Figure 20: An example of a Petri Net*

**What does it do well:** PN's are intuitive and model concurrency (partial order) well. PNs have been adapted to supporting temporal events and probabilities by different extensions. By supporting these extensions this becomes a very useful modeling tool.

**Shortcomings:** The disadvantages of a PN are that the size of the nets for modelling very complex systems become difficult to validate because the number of reachable markings blows up, making it analytically intractable. In other words, PNs are considered state space models that allow complex interactions at the system and sub-system levels but are prone to the state space explosion problem. The state space explosion problem is the size of a state space with respect to the structural scale of a PN and has a tendency of growing exponentially.

The next technique is different from the others with regard to using probabilities and undirected paths for inference. BN provides the capability to reason about the domain once a joint distribution is constructed that includes prior evidence.

**What is it:** BN are Probabilistic Graphical Models (PGM) that represent causality inference using variable conditional dependence. A DBN extends a BN by relating variables to each other over adjacent temporal steps. BNs use the underlying Bayes Theorem as explained below that is followed by an example.

**How does it work:** To propagate the level of belief in a hypothesis that is put to the network, that level of belief can be formulated to indicate the level of trust that is placed in a system based on the evidence that supports (or negates) the hypothesis. For example, from Bayes theorem:

$P(X|Y) = (P(Y|X) * P(X)) / P(Y)$, which is the Posterior = (Likelihood * Prior) / Evidence). Posterior is after an observation has occurred, Likelihood is the probability that the event will happen, and Prior is the data before it is observed/evidence set where something is known.

The following example, shown in Figure 22, applies a Bayesian analysis to two different boards, the one on the left is a low-cost hobbyist processor board with an Arduino-like processor (overseas manufacturer and open source software) versus a TI MSP430 development board (US manufacturer, reputable suppliers, etc.). The microprocessor trust metric and supplier trust metric nodes have a causal relationship with the board. The conditional probability table is constructed by the first column being hardware trust metric using values (none, low, low to medium, medium to high and high), while the second column is the supplier trust metric with values for each row (unsatisfactory, marginal, satisfactory, and excellent). The probabilities are shown in the next two columns corresponding to the different processors. By intuition, the TI MSP430 provides a higher trust level than the Arduino-like chip, since the manufacturer is US-based, the supplier evaluation ratings and its hardware design integrity are higher as well. These were compared to the Arduino-like processor where the supplier did not have a rating and did not have complete design artifacts (back to the hardware trust metric for (Logical Equivalence, Signal Activity Rate, Structural Architecture, Functional Correctness, and Power Consumption). By setting evidence on a node, one can reason about the outcome in a downward path, so setting the supply chain node to unsatisfactory will cause the board node to change by observing its state. In the case of the evidence being set on the bottom node, the two top nodes will change accordingly. This effect of change when evidence is set is known as inference and BNs provide this capability.

**What does it do well:** Bayesian Networks are able to perform casual inference and reason with uncertainty. When reasoning with uncertainty, the BN model can update the posterior probabilities of other states of the system when evidence is set. This technique is well suited for autonomous robotics system because of their nature in uncertainty and having the capability to reason about new evidence gained from sensors about tasks or environment.

**Shortcomings:** There is no standard method for constructing a BN. However, there are techniques for determining if the model has been constructed correctly.

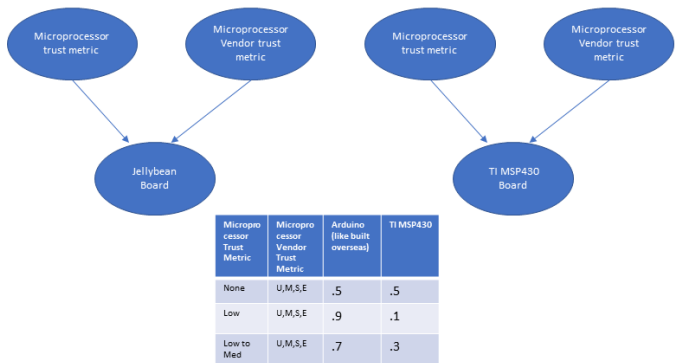

*Figure 22: A BN example of two microprocessors*

Each of these techniques has both pros and cons associated with them as enumerated above. There have been three prongs to the development of these techniques. The first prong being the ongoing work to extend trees and PN by increasing their capabilities after they were first introduced in 1960's. The second prong, being a combined approach, where pre-preprocessing is done in (trees and PN) and post-processing is completed in BNs. The third prong being an only BN usage model for security analysis.

In addition to the approaches described above, we also observed that attack graphs combined with CVSS base metrics were common methods for security assessment. Others have used PNs to cover a small portion of the architecture, while others are using AI techniques to perform prediction on security outcomes. A common problem with these techniques is that they do not support uncertainties that may arise from autonomous robotic systems.

On the other hand, BNs provide a number of benefits over these other approaches. These include the casual inference where reasoning can be performed from observations and the compactness of information can be encoded into joint probability distributions. In fact, the efficiency of inference algorithms can be seen as main reason to the success of BNs, since querying general graphs is an NP-hard problem [78], [79], [80]. Another strength of BNs is their ability to update the model, i.e., compute a posteriori distribution, when new information is available [80]. Using a BN with the associate metrics is one potential assessment solution for autonomous robotic systems.

IX. A SOLUTION OUTLINE

We outline a solution that takes a number of the autonomous robotic system layers and brings them together to form a holistic trust model. This trust model is different from other approaches because we are proposing a complete solution from a security perspective, eliminating the gaps left by other techniques in the evaluation of system, hardware, and software layers. Our model also includes AI robustness attributes and supply chain characteristics. The overall complexity of the space and the security problems are bad enough in a controlled environment, now add high-value targets in an unconstrained environment where they get much worse. The unconstrained environment makes the problem computationally intractable using earlier approaches. Thus, there is a need to come up with a way to make assessment more computationally feasible.

A probabilistic approach to analysis using BNs provides a natural way to reason about uncertainty. BN-based models allow for efficient factorization of the set of system states, without the need for an explicit representation of the whole joint distribution; moreover, they have the additional advantage of inference algorithms available for the analysis of any posteriori situation of interest (i.e. evidence can be gathered by a monitoring system) [81]. Bayesian Inference simplifies the way to reason about a complex domain problem like security for autonomous robotic systems. Our survey findings did reveal some usage of BNs, but these researchers did not define address complete systems, nor did they include cost values (collateral damage and perceived target value) in their work. Our goal is to expand this work and apply it to more complex autonomous systems.

In order to represent an autonomous robotic system architecture and assess the security of it, we have discussed the different layers (system, hardware, software, Cognitive/AI, and supplier chain) as independent trust metrics. These individual trust metrics account for the different levels depending on the security features supported by the autonomous robotic system and may be expanded to cover other elements of importance.

With each individual trust metric and its associated level, we also need to account for the collateral damage that may result from an attack on that part of the system as well as the perceived target value of that element. In other words, we need to account for the adversary's actions and by combining these values with the trust metric we get the general probability equation:

$TM = LV*AER* AED *ATA$, where TM = trust metric, LV = level value, AER = probability of adversary exploit reward, AED = probability of adversary exploit damage, and ATA = likelihood of an adversary taking action to exploit.

The combination of these values provides a set of metrics that can be assigned to each corresponding component in the system. The BN will provide the casual inference by linking these components and the values for the conditional probability tables. A full joint distribution is defined as the product of the conditional distribution of each node. This is shown in the equation below where the left-hand side is the joint distribution, the center is the conditional probability (using the chain rule), and the right-hand side is the conditional probability given the parents.

$$P(x_1 \dots x_n) = \prod_{i-1}^{n} P(x_i | x_1 \dots x_{(i-1)}) = \prod_{i-1}^{n} P(x_i | \text{parents}(x_i))$$

In general, a trust metric for a given level is created as a function, $f_i$, where $f_i$ = Factor 1 * Factor 2 * … * Factor N. The number of metric values is a function of the granularity of the analysis, and generally ranges from 3 to 7 with 5 being a reasonable tradeoff between resolution and complexity. From these factors, a trust metric, T, is derived by normalizing the $f_i$ over the range, so $T = 0 \leq |f_i| \leq 1$.

For example, at the system level trust metric, we start with the related CC's EAL 1 to 5 levels and combine them with cost/reward/likelihood values, this provides five levels for the system metric and each having a cost/reward/likelihood value, then that is normalized.

$ST_n = EAL_n * AER_n * AED_n * ATA_n$, where n is the level
ST = system trust metrics ranging in a discretized continuous set from [0, 1] where 1 is the highest and zero the lowest trust set. Where five levels will be defined for the range. The other layers HW, SW, AI and Supply chain follow the same pattern.

Next, we take the hardware design integrity values from the hardware evaluation section and do the same for the cost/reward/likelihood values.

HWT = hardware trust metrics will have a range [0, 1].

Next, we take the software base metric in the software evaluation section and do the same with the cost/reward/likelihood values.

SWT = software trust metrics will have a range [0,1].

Since cognitive/AI is a new area and using the distance metric seems to fall into the same thought process as the others, we come up with five levels with a range between [0,1] in which the closer to the certified area signifies greater protection/risk than moving further away where perturbation is easier to detect.

AR = AI robustness trust metrics will consist of five levels of distance error + cost values ranging in a discretized continues set from [0, 1] where ¬ [0,1], so that 0 is the lowest and 1 becomes the highest trust metric level.

Next, we take the supplier chain metric in the supply chain evaluation section and do the same with the cost/reward/likelihood values.

VT = Supply chain trust metrics will have a range [0,1].

Finally, we take these individual trust metrics that represent the system levels and sum the different parts into a whole system trust metric.

The following equation represents the joint distribution over all variables, where $P(X_i | P_a(X_i))$ is the conditional probability distribution (CPD) for each variable in the network.

$$P(ST, HWT, SWT, AR, VT) = \prod_i p(X_i | P_a(X_i))$$

By using the above equation, we can reason about outcomes of the network while making observations. This allows the intractable problem to be computationally feasible using Bayesian Inference. BNs fulfill the local Markov property, so that each variable is conditionally independent of its non-descendants given its parent variables. This property reduces the joint probability to a compact form by using the chain rule.

X. CONCLUSION

We surveyed the trust metric space to determine if a set of security metrics were well defined and covered a complete robotic system (the results were in Section II). A mind map of a robot system broke down the different layers into system, hardware, software, cognitive layer, and supplier chain to provide a holistic security view. Our findings showed that system level trust metrics are difficult and complex, and several research papers scaled the problem to a small set of components or just a specific area of a system. Table 3 is a summary of the findings that cover a holistic system trust model.

*Table 4: Summary of Trust Metrics*

| Trust Level | Trust Metric Recommendation | Values | Comment |
|---|---|---|---|
| System | EAL 1 to 5 | [0,1] | Security and Safety |
| Hardware | Hardware Component | [0,1] | |
| Software | Base Impact | [0,1] | |
| AI Robustness | Distance Metric | [0,1] | |
| Supply Chain | NG Scorecard | [0,1] | |

We believe that using the Bayes Inference is the correct choice to build on, since it provides several benefits to overcome uncertainties for a complex system like an autonomous robotic system. By using Bayesian Inference, we also remove the intractable problem to a computational feasible one. We provide a brief set of definitions for each of the CPD nodes that will represent the joint distribution in a BN.

As autonomous robotic systems become more popular, the standards that govern them should consider the usage of trust metrics for security. Several standards like the IEEE global initiative on Ethics and the EU's regulatory work on ethics are targeting the bias/transparence for AI systems, but little is being said about the security of these systems. It is our belief that this survey will help establish a set of security metrics that the ethics groups can utilize. Another consideration for security trust metrics is the autonomous vehicle space.

The industry is seeing a movement in the autonomous vehicle space where the Society of Automotive Engineers (SAE) has defined a scale for automation. Level 0 is fully human controlled, and level 5 is fully automated. We are witnessing the gradual paradigm shift to more technology being incorporated into the vehicle as the goal of level 5 is being realized. Even with these advancements there have been several accidents where life was lost/injured. As these systems shift to more AI capabilities, this will increase the attack vector for nefarious actors.

Future research work is to define these trust metrics that can be used in a model for security analysis within the autonomous robotic system.